\documentclass[letterpaper]{article} 
\usepackage{aaai2026}  
\usepackage{times}  
\usepackage{helvet}  
\usepackage{courier}  
\usepackage[hyphens]{url}  
\usepackage{graphicx} 
\usepackage{natbib}  
\usepackage{caption} 
\usepackage{algorithm}
\usepackage{algorithmic}

\usepackage[table]{xcolor}
\usepackage{makecell}
\usepackage{multirow} 
\usepackage{makecell}

\usepackage{amsmath}
\usepackage{amsfonts}
\usepackage{booktabs}
\usepackage{newfloat}
\usepackage{listings}
\DeclareCaptionStyle{ruled}{labelfont=normalfont,labelsep=colon,strut=off} 
\floatstyle{ruled}
\newfloat{listing}{tb}{lst}{}
\floatname{listing}{Listing}
\title{PathFLIP: Fine-grained Language-Image Pretraining for Versatile \\ Computational Pathology}

\author{
    Fengchun Liu\equalcontrib\textsuperscript{\rm 1},
    Songhan Jiang\equalcontrib\textsuperscript{\rm 1},
    Linghan Cai\textsuperscript{\rm 1},
    Ziyue Wang\textsuperscript{\rm 2},
    Yongbing Zhang\textsuperscript{\rm 1}
}
\affiliations{
    \textsuperscript{\rm 1}Harbin Institute of Technology, Shenzhen, School of Computer Science and Technology \\
    \textsuperscript{\rm 2}National University of Singapore, Department of Electronic and Computer Engineering\\
    \{24s051033, 23s151019, cailh\}@stu.hit.edu.cn, e1374378@u.nus.edu, ybzhang08@hit.edu.cn
}

\usepackage{bibentry}

\begin{document}

\maketitle

\begin{abstract}
While Vision-Language Models (VLMs) have achieved notable progress in computational pathology (CPath), the gigapixel scale and spatial heterogeneity of Whole Slide Images (WSIs) continue to pose challenges for multimodal understanding.
Existing alignment methods struggle to capture fine-grained correspondences between textual descriptions and visual cues across thousands of patches from a slide, compromising their performance on downstream tasks.
In this paper, we propose PathFLIP (\textbf{Path}ology \textbf{F}ine-grained \textbf{L}anguage-\textbf{I}mage \textbf{P}retraining), a novel framework for holistic WSI interpretation.
PathFLIP decomposes slide-level captions into region-level subcaptions and generates text-conditioned region embeddings to facilitate precise visual-language grounding.
By harnessing Large Language Models (LLMs), PathFLIP can seamlessly follow diverse clinical instructions and adapt to varied diagnostic contexts.
Furthermore, it exhibits versatile capabilities across multiple paradigms, efficiently handling slide-level classification and retrieval, fine-grained lesion localization, and instruction following. 
Extensive experiments demonstrate that PathFLIP outperforms existing large-scale pathological VLMs on four representative benchmarks while requiring significantly less training data, paving the way for fine-grained, instruction-aware WSI interpretation in clinical practice.
\end{abstract}

\begin{links}
    \link{Code}{https://github.com/cyclexfy/PathFLIP}
\end{links}

\begin{figure}[t]
\centering
\includegraphics[width=0.9\linewidth]{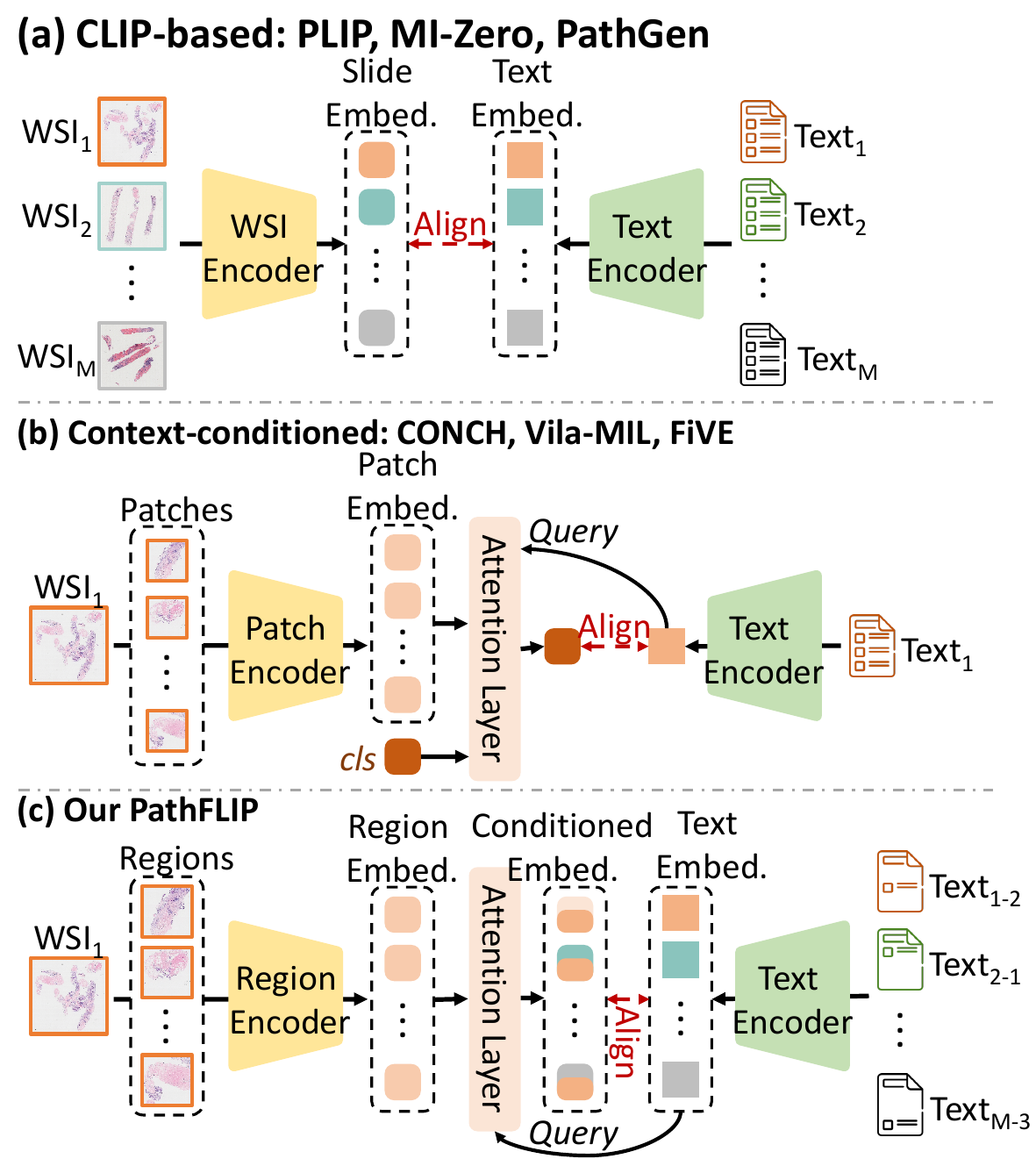}
\caption{   
{Comparison of PathFLIP with previous methods in vision-language pathology modeling.}
(a) CLIP-based methods perform coarse global feature alignment between images and text.
(b) Context-conditioned approaches use textual cues to guide attention but focus on global alignment.
(c) Our PathFLIP enables fine-grained pathology analysis by aligning localized image embeddings with semantically matched text segments.}
\label{motivation}
\end{figure}

\section{Introduction}

Whole Slide Images (WSIs) are fundamental to pathology diagnosis\cite{ cai2025attrimil}, yet their gigapixel resolution and complex spatial structures pose significant challenges for multimodal learning\cite{jiang2024multimodal,jiang2025uncertainty}. Since each slide contains thousands of heterogeneous patches, achieving fine-grained visual-textual alignment is challenging. To address this, the emergence of large-scale multimodal datasets \cite{Slidechat}, pairing WSIs with expert-authored slide captions and driving progress in the field. These captions are extended and detail-rich, reflecting clinical workflows and supporting diagnostic reasoning.

This expanding multimodal opportunity aligns with advances in Vision-Language Models (VLMs), particularly those leveraging contrastive pretraining paradigms exemplified by CLIP \cite{CLIP, Dreamlip, xiao2025flair}, which have shown impressive capabilities in learning joint image-text representations across general domains.
However, when applied to computational pathology (CPath), these models \cite{MI_Zero, PLIP} face unique challenges arising from the gigapixel scale and spatial heterogeneity of WSIs (Figure~\ref{motivation}(a)). Most existing methods extract global feature of a slide from thousands of patches, followed by global alignment with slide-level captions. Nevertheless, this coarse-grained strategy fails to capture fine-grained correspondences between visual regions and textual descriptions \cite{Pathgen}.
To overcome these limitations, some approaches adapt VLMs to the multiple instance learning (MIL) framework, aggregating patch-level features into slide-level representations through attention mechanisms guided by textual cues \cite{Vila-mil, Pathalign, tang2025ipgphormer, dong2025disentangled}, as shown in Figure~\ref{motivation}(b). These methods leverage contextual signals from captions to highlight semantically relevant regions before performing slide-text alignment between visual and textual embeddings \cite{Focus, FIVE}.
While these strategies enhance localization, they typically depend on weak or implicit region-text supervision and fall short of fully emulating the diagnostic reasoning of pathologists, which requires grounding from region-level observations to slide-level conclusions \cite{Pathfinder}.

Targeting the above challenges, we propose PathFLIP (\textbf{Path}ology \textbf{F}ine-grained \textbf{L}anguage-Image \textbf{P}retraining), a novel framework for fine-grained WSI analysis.
As illustrated in Figure~\ref{motivation}(c), PathFLIP establishes region-text correspondences by aligning localized visual features with semantically matched text segments.
Concretely, we design a Region Q-Former to extract region-level embeddings from the WSI. Meanwhile, the slide-level captions are decomposed into region-level subcaptions that reflect localized pathological semantics and serve as anchors to guide region-level grounding, promoting the disentanglement of interpretable clinical concepts.
At the region level, we introduce a self-supervised contrastive objective that treats region-subcaption pairs from the same slide as positives, while considering all other combinations in the batch as negatives. This encourages precise and discriminative region-text alignment.
To retain a global understanding, we align WSI representations with their corresponding full captions using a global contrastive loss.
This pretraining strategy enhances alignment accuracy and interpretability by enabling fine-grained supervision without manual regional annotations.

Building upon these foundations, PathFLIP demonstrates exceptional capabilities, seamlessly adapting across diverse multimodal pathology tasks.
Through efficient integration with Large Language Models (LLMs), PathFLIP acquires advanced instruction-following abilities such as caption generation and visual question answering (VQA) within a unified framework.
At the region level, PathFLIP excels at semantic understanding, accurately identifying lesion-relevant areas within WSIs. This enables detailed region-level visual grounding of pathological descriptions.
For slide-level tasks, PathFLIP delivers robust performance in zero-shot classification and retrieval, achieving competitive results and significantly surpassing previous large-scale VLMs across four benchmarks.
Collectively, these strengths position PathFLIP as a versatile, unified model for computational pathology, capable of effectively addressing various multimodal understanding challenges.
Key contributions are as follows:
\begin{itemize}
    \item We propose PathFLIP, a fine-grained language-image pretraining framework specifically for WSIs. By leveraging subcaption–guided embedding alignment, PathFLIP enables accurate region-level grounding without expert annotations. These localized semantics are further aggregated to support holistic slide-level diagnosis and comprehensive multimodal interpretation, closely aligning with the typical workflow of pathologists.
    \item By integrating LLMs, PathFLIP gains the flexibility to address a wide range of multimodal pathology tasks at both the region and slide levels, including lesion localization, instruction-following, and pathological diagnosis.
    \item Extensive experiments demonstrate that PathFLIP consistently outperforms existing large-scale VLMs on diverse clinical benchmarks, requiring significantly less training data. Its interpretable and multimodal capabilities enable practical deployment in diagnostic support and clinical AI applications.
\end{itemize}

\section{Related Work}

\subsection{Pretraining in Computational Pathology}
Pretraining in computational pathology leverages large-scale datasets and self-supervised learning to produce general-purpose visual representation models \cite{pathsurvey}.
Unimodal models focus solely on visual features, utilizing hierarchical structures in WSIs through pyramid Vision Transformers \cite{HIPT} and masked autoencoding with self-distillation \cite{UNI, Prov-GigaPath}.
Recent works bridge the gap between visual and textual modalities by employing cross-modal contrastive learning \cite{CHIEF, Cpath-omni}.
PathGen \cite{Pathgen} retrieves semantically relevant patches using cross-modal signals to train a high-quality patch encoder.
TITAN \cite{TITAN} aligns regional embeddings with region captions and slide-level reports, while CONCH \cite{CONCH} enables fine-grained alignment via context-aware patch embeddings.
In contrast, our PathFLIP unifies slide-level and fine-grained alignment within a single framework, achieving superior performance with significantly fewer training samples.

\subsection{Pathological Vision-Language Models}
While CLIP-based methods in computational pathology \cite{MI_Zero, PLIP, Pathgen, liu2025prototype,yuan2018unsupervised} apply contrastive learning to align slide representations with their global captions, they lack supervision signals for fine-grained region-text alignment.
Motivated by the success of LLMs, several studies have explored more effective ways to bridge pathology images and textual semantics.
For example,
PathAlign \cite{Pathalign} extends the BLIP framework with LLMs to support diverse WSI applications.
SlideChat \cite{Slidechat} utilizes a slide encoder to obtain global slide embeddings, which are sent into an LLM for downstream slide-level multimodal tasks.
PRISM2 \cite{PRISM2} introduces a two-stage training framework that aligns slide embeddings with diagnostic summaries, and then incorporates LLM-guided supervision using clinical prompts.
Nevertheless, existing methods fail to achieve fine-grained reasoning and ground visual evidence based on textual cues.
In contrast, our approach unifies multimodal understanding across regions and slide levels, while supporting diverse pathology tasks in real-world clinical settings.

\begin{figure*}[t]
\centering
\includegraphics[width=0.95\textwidth]{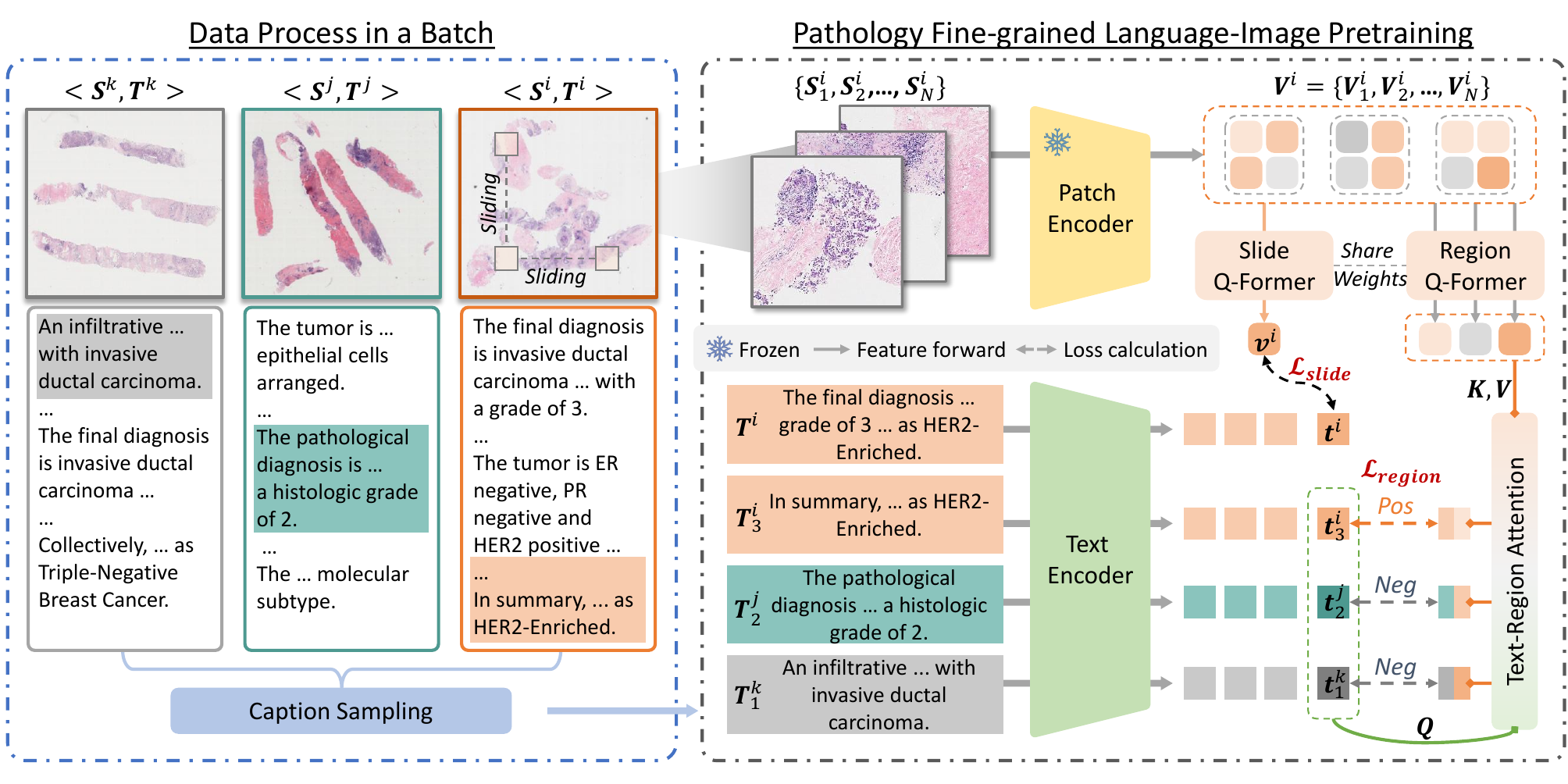}
\caption{
{Overview of PathFLIP.} Given a slide-caption pair \textless$S^i$, $T^i$\textgreater, the slide $S^i$ is divided into $N$ regions $\{S^{i}_{1}, \ldots, S^{i}_{N}\}$. 
We use Slide Q-Former and Region Q-Former to extract slide-level and region-level features. Captions $\{T^k, T^j, T^i\}$ are decomposed and sampled to obtain region-level subcaptions $\{T^{k}_1, T^{j}_2, T^{i}_3\}$. The slide-level contrastive loss $\mathcal{L}_{slide}$ aligns the global image feature $v^i$ with its corresponding text feature $t^i$. The region-level contrastive loss $\mathcal{L}_{region}$ encourages alignment between region-image and subcaption pairs from the same slide as positive pairs, while treating all others in the batch as negatives.
}
\label{overview}
\end{figure*}

\section{Methodology}

\subsection{Overall Structure of PathFLIP}
\begin{figure*}[t]
\centering
\includegraphics[width=0.90\linewidth]{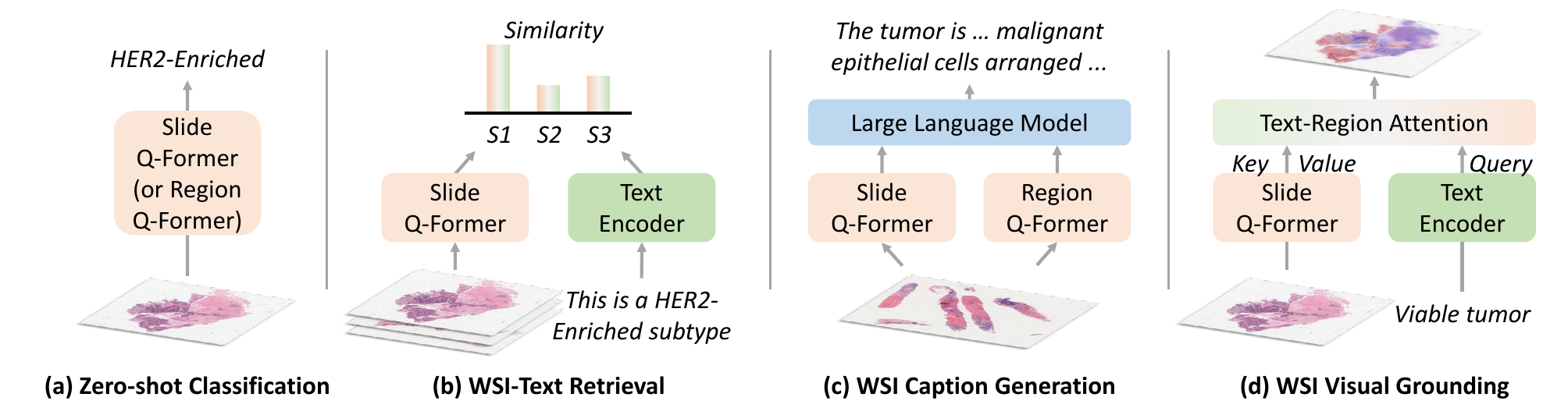}
\caption{
{PathFLIP serves as a versatile tool in computational pathology.} It accommodates a diverse range of multimodal pathology tasks at both slide and region levels. In (b), ``$Si$'' refers to the similarity between the $i$-th slide and the input text.
}
\label{downstream}
\end{figure*}

The overall framework of PathFLIP is illustrated in Figure~\ref{overview}.
We describe the visual feature extraction process at both slide and region levels, then construct region-level subcaptions and their textual representations.
We then introduce a contrastive pretraining objective that jointly aligns visual and textual features, incorporating both text-conditioned region-level alignment and global slide-level alignment.
This framework enables to learn semantically aligned visual representations, facilitating effective transfer to various downstream tasks in Figure~\ref{downstream}. 

\subsection{Visual Representation}

\subsubsection{Region-level Visual Feature Extraction}

Given the gigapixel resolution of WSIs, we adopt a two-stage partitioning strategy to extract region-level visual features from an input slide $S^i$.
First, the slide is divided into $N$ non-overlapping regions $\left\{ S^{i}_n \right\}_{n=1}^{N}$ using a sliding window.
Each region $S^{i}_n$ is further partitioned into $M$ non-overlapping patches, and each patch is encoded into a $d$-dimensional feature vector using a pretrained pathology-specific visual encoder.
This results in a region-level feature matrix $V^{i}_n \in \mathbb{R}^{M \times d}$ for each region $n$ of slide $S^i$.
The slide-level feature matrix is obtained by concatenating all patch-level features across all regions, denoted as $V^i \in \mathbb{R}^{MN \times d}$.

\subsubsection{Region and Slide Q-Former}
After feature extraction, we obtain $N$ non-overlapping regions per slide, each consisting of $M$ independently encoded patch features. Due to the lack of spatial dependency modeling and redundancy among patch representations, compact and informative semantic aggregation becomes essential.
Inspired by BLIP-2 \cite{li2023blip}, we adopt a Q-Former module to enhance intra-region semantic fusion. It consists of self-attention, cross-attention, and feed-forward layers, and uses a learnable set of semantic query tokens $Q \in \mathbb{R}^{N_q \times d}$. Since $N_q \ll M$, the Q-Former efficiently distills the salient semantics.

We apply the Region Q-Former $f_{rv}(\cdot,\cdot)$ to each region feature $V^i_{n}$ to obtain fine-grained embeddings:
\begin{equation}\label{eq3}
    v^i_{n} = f_{rv}(Q, V^i_{n}) \ ,
\end{equation}
where $v^i_{n} \in \mathbb{R}^{N_q \times d}$ represents the semantic embedding of region $n$ in slide $S^i$.
The resulting region embeddings $\{ v^i_{n} \}_{n=1}^{N}$ are concatenated along the first dimension to form $\textbf{v}^{\text{reg}} \in \mathbb{R}^{N N_q \times d}$.
Similarly, the Slide Q-Former $f_{sv}(\cdot,\cdot)$ is applied to the full patch set $V^i$ of slide $S^i$ to derive a global representation:
\begin{equation}\label{eq2}
    v^i = f_{sv}(Q, V^i) \ ,
\end{equation}
where $v^i \in \mathbb{R}^{N_q \times d}$ denotes the holistic embedding of the slide.
The Q-Former shares weights at the region and slide levels to ensure consistency and efficiency in feature aggregation across scales. The resulting global embedding $v^i$ and regional features $\textbf{v}^{\text{reg}}$ collaboratively support holistic slide understanding and fine-grained vision-language alignment.

\subsection{Text Representation} 
The pretraining data used for our PathFLIP follows SlideBench \cite{Slidechat}, which provides slide-level captions containing long and detailed descriptions.
These captions reflect the standard clinical workflow of identifying ROIs, conducting diagnostic evaluations, and integrating findings into a slide-level conclusion.
To align the fine-grained textual descriptions with the regional visual features set $\textbf{v}^\text{reg}$ extracted, we construct a corresponding set of fine-grained textual representations.
Specifically, each slide-level caption is first segmented into individual sentences. Then, we construct $K$ subtext segments $\left\{ T^i_{k} \right\}_{k=1}^{K}$ by randomly sampling subsets of sentences from the original caption $T^i$ using a probabilistic strategy.
Each subcaption is subsequently encoded into a fixed-dimensional representation $t^i_{k} \in \mathbb{R}^{d}$ using a learnable text encoder $f_t(\cdot, \cdot)$:
\begin{equation}\label{eq5}
t^i_{k} = f_t (\text{[CLS]}, T^i_{k})\ .
\end{equation}
The complete set of region-level textual features $\{ t^i_{k} \}_{k=1}^{K}$ is defined as $ \textbf{t}^\text{sub} \in \mathbb{R}^{K \times d}$.
We preserve the full-text caption and encode it into a global textual representation $t^i \in \mathbb{R}^{d}$ as:
\begin{equation}\label{eq4}
t^i = f_t (\text{[CLS]}, T^i) \ .
\end{equation}
During training, subcaptions are randomly sampled for region-level alignment, while the full caption supports global representation learning.

\subsection{Vision-Langugae Alignment}

\subsubsection{Text-conditioned Region Feature Alignment}

To capture fine-grained pathological semantics, it is necessary to establish semantic associations between the regional feature set $\textbf{v}^\text{reg}$ and the subcaption feature set $\textbf{t}^\text{sub}$, thereby enabling region-level alignment between WSI features and text representations. To this end, we employ a cross-attention mechanism, where subcaption features are used as queries to attend to regional visual features. This process is shown as:
\begin{equation}\label{eq9}
\textbf{v}^\text{tc} = \text{softmax}\left(\frac{W_q \textbf{t}^\text{sub} ( {W_k \textbf{v}^\text{reg}})^\top}{\sqrt{d}}\right) W_v \textbf{v}^\text{reg} \ ,
\end{equation}
where $\textbf{v}^\text{tc}$ denotes the text-conditioned query results of attended features and $W_q$, $W_k$, and $W_v$ are learnable projection matrices.

To enhance region-text alignment, we incorporate subcaption features from other image-text pairs within the same batch as negative queries. For a given image-text pair $\langle S^i, T^i \rangle$, the subcaptions of $T^i$ serve as positive conditions, while those from other texts $T^j$ $(j \neq i)$ act as negatives for contrastive learning.
To reduce memory and computational cost, we randomly sample one subcaption from each of the other $B-1$ pairs, rather than using all. Consequently, the total number of text-conditioned queries becomes $L = K + B - 1$, where $K$ is the number of subcaptions for $T^i$ and $B$ is the batch size.
We adopt the LogSigmoid loss for regional contrastive learning:
\begin{equation}\label{eq10} 
\mathcal{L}_{region} = \frac{1}{L}\sum_{l=1}^{L} -\text{log}( \frac{1}{1 + \text{exp}(y_{l} \cdot -\eta < \textbf{v}_{l}^\text{tc}, \textbf{t}_{l}^\text{sub}>)} )  \ ,
\end{equation}
where $y_l = +1$ for positive pairs and $-1$ for negative ones, and $\eta$ is a learnable temperature parameter. The similarity $<\textbf{v}_{l}^\text{tc}, \textbf{t}_{l}^\text{sub}>$ between the text-conditioned visual query result $\textbf{v}_l^{\text{tc}}$ and the textual query $\textbf{t}_l^{\text{sub}}$ is computed via dot product.
Final region-level alignment is obtained by mean pooling over the $L$ query results.
This self-supervised formulation encourages alignment of semantically related regions and subcaptions, which we find essential for fine-grained visual-text grounding.

\subsubsection{Global Feature Alignment}
We adopt a contrastive learning strategy to align global-level semantics of vision and language modalities. Within each training batch, matched image-text pairs are treated as positive samples, while mismatched pairs are treated as negatives. We compute the cosine similarity between global image embeddings $v^{i}$ and text representations $t^{i}$.
We average the $N_q$ query features in $v^i$ to obtain $\bar{v}^i$ for image-text similarity computation.
The contrastive loss is formulated as:
\begin{equation}\label{eq6} 
\mathcal{L}_{i2t}=
-\text{log}(\frac{\text{exp}(<\bar{v}^i,t^i>/\tau)}
{\Sigma_{b=1}^{B}\text{exp}(<\bar{v}^i,t^b>/\tau)}) \ ,
\end{equation}
\begin{equation}\label{eq7} 
\mathcal{L}_{t2i}=
-\text{log}(\frac{\text{exp}(<\bar{v}^i,t^i>/\tau)}
{\Sigma_{b=1}^{B}\text{exp}(<\bar{v}^b,t^i>/\tau)}) \ ,
\end{equation}
where $B$ is the batch size, $<\cdot,\cdot>$ represents the cosine similarity calculation and $\tau$ is the temperature hyperparameter. The total global contrastive loss is: 
\begin{equation}\label{eq8} 
\mathcal{L}_{slide} = \frac{1}{2}(\mathcal{L}_{t2i}+\mathcal{L}_{i2t}) \ .
\end{equation}

Our final loss of vision-language alignment is the sum of global loss $\mathcal{L}_{slide}$ and regional loss $\mathcal{L}_{region}$:
\begin{equation}\label{eq11}
    \mathcal{L} = \mathcal{L}_{region} + \mathcal{L}_{slide}.
\end{equation}

\subsection{Transfer for Visual Understanding}

\subsubsection{Supervised Instruction Fine-tuning}
Building upon the pretrained model, we fine-tune it by integrating with an LLM to enable instruction-following for tasks such as captioning and question answering.
Specifically, we concatenate the slide-level embedding $v^i$ with the set of region-level embeddings $\{v^i_{n}\}_{n=1}^{N}$ and project them into the LLM’s embedding space using a linear transformation $\sigma(\cdot)$. A task-specific prompt embedding $t_p$ (e.g., for caption generation) is prepended to the projected visual features, which are then fed into the LLM for autoregressive text generation.
The entire model is optimised using a language modeling objective:
\begin{equation}\label{eq12}
\mathcal{L}_{lm} = \sum_{j=1}^{L_t} -\log P_{lm}({\mathcal{T}}^{i}_j \mid \mathcal{{T}}^{i}_{<j}, 
\sigma([v^i, v^i_1, \ldots, v^i_N]), t_p),
\end{equation}
where $L_t$ denotes the length of the generated sequence, $P_{lm}$ is the token-level conditional distribution predicted by the LLM, ${\mathcal{T}^{i}_j}$ indicates the $j$-th ground truth token, and $\mathcal{T}^{i}_{<j}$ represents the preceding tokens in the sequence.
The model acquires enhanced multimodal comprehension and effective instruction-following abilities by integrating visual features with the LLM in this unified framework.

\begin{figure}[t]
\centering
\includegraphics[width=0.95\linewidth]{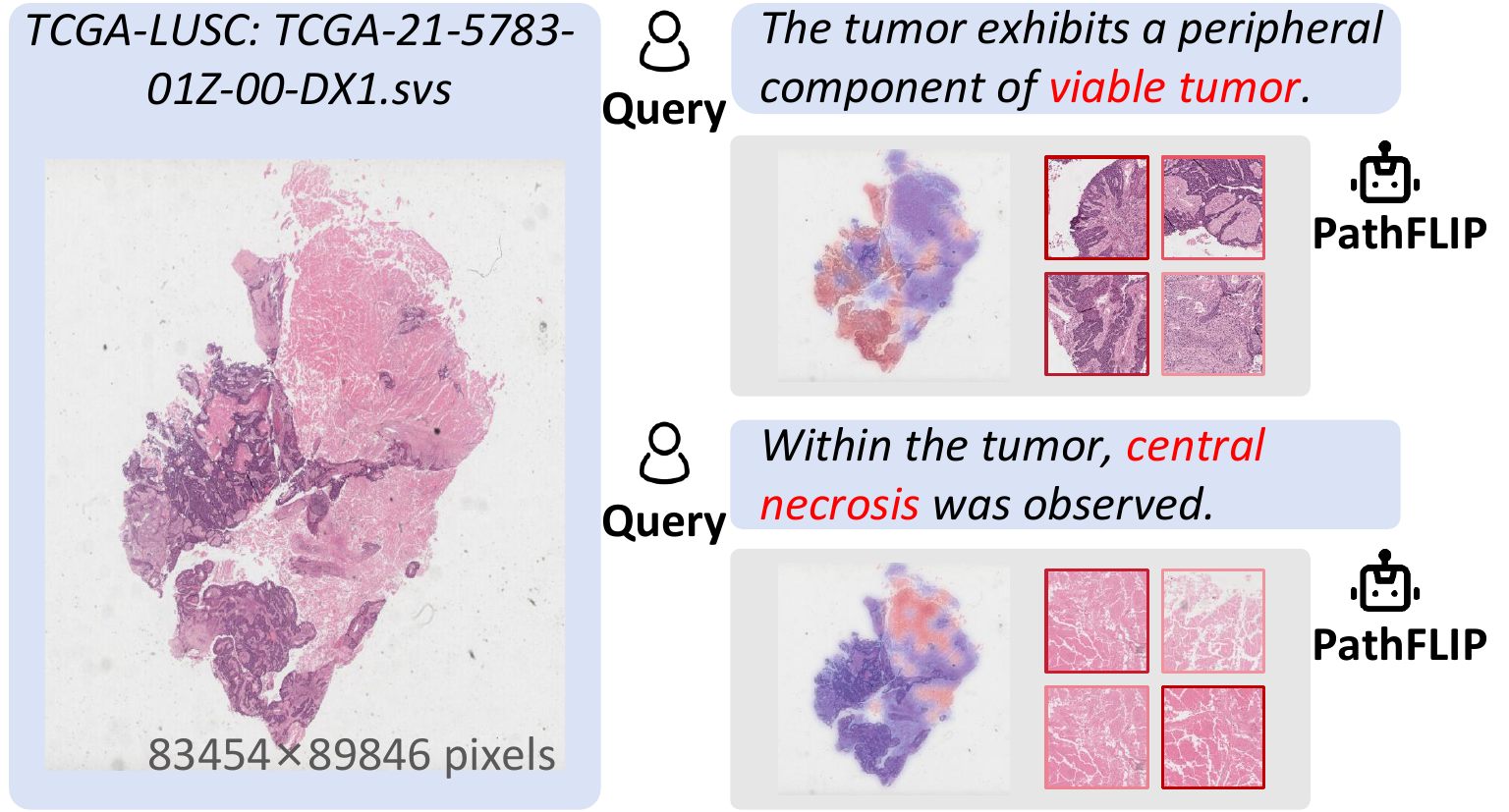}
\caption{
{Visual grounding results}. High-attention areas are highlighted in red in the heatmap, with red boxes marking the corresponding regions.
}
\label{grounding}
\end{figure}

\subsubsection{Generalizable Visual-Language Understanding}
We project pathology images and textual descriptions into a shared embedding space, where semantically aligned image-text pairs are pulled closer via contrastive learning.
For an input slide-caption pair $\langle S^i, T^j \rangle$, we extract their embeddings $v^i$ and $t^j$, and compute the similarity score.
Moreover, our Region Q-Former module outputs region-level embeddings $v^{i}_n$, enabling fine-grained visual-language understanding.

\subsubsection{Visual Grounding}
Our method aligns with the diagnostic workflow of pathologists by integrating regional observations into coherent narratives to support downstream interpretation and visual grounding.
For each slide, region-level representations are extracted using the Region Q-Former. A text-region attention layer is then applied to compute attention scores between each region and its corresponding subcaption, enabling precise region-level alignment and grounding.
As shown in Figure~\ref{grounding}, we visualize the attention maps corresponding to two subcaptions from a given slide. PathFLIP successfully localizes regions of interest such as ``viable tumor'' and ``central necrosis.''
These results demonstrate that our framework enables fine-grained visual grounding guided by natural language prompts, facilitating transferable reasoning and clinical localization. Notably, the grounding ability is unique to PathFLIP and is not available in most previous approaches. The detailed analyses of the key feature are in supplementary materials.


\begin{figure*}[t]
\centering
\includegraphics[width=1.0\linewidth]{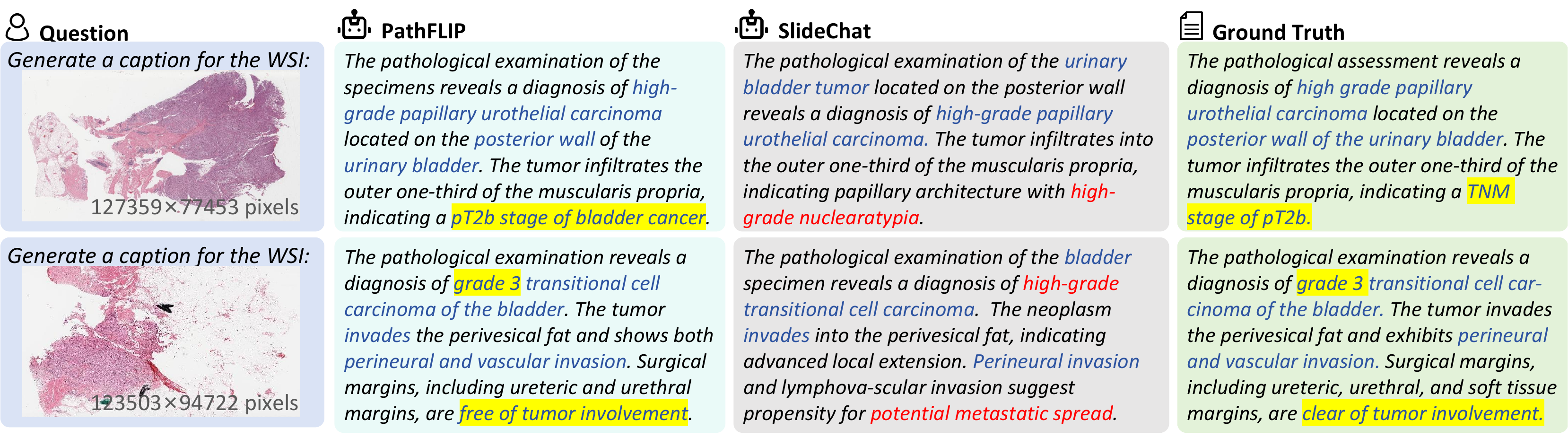}
\caption{{Caption generation comparison}. Blue indicates correct matches, red indicates incorrect or imprecise matches, and yellow backgrounds emphasize important information matches.}
\label{caption}
\end{figure*}

\begin{table*}
\centering

\resizebox{\textwidth}{!}{%
\begin{tabular}{l lll lll lll lll c}
\toprule
 AUC & \multicolumn{3}{c}{BRCA} & \multicolumn{3}{c}{COAD} & \multicolumn{3}{c}{LUAD} & \multicolumn{3}{c}{LSCC} & \multirow{2}{*}{Average} \\
\cmidrule(lr){2-4}
\cmidrule(lr){5-7}
\cmidrule(lr){8-10}
\cmidrule(lr){11-13}
 Model & \textit{PIK3CA} & \textit{MAP3K1} & \textit{GATA3} & \textit{KRAS} & \textit{PIK3CA} & \textit{TP53} & \textit{TP53} & \textit{STK11} & \textit{KRAS} & \textit{TP53} & \textit{PIK3R1} & \textit{KEAP1} & \\
\midrule
MI-Zero \cite{MI_Zero}  & 0.5321 & 0.5542 & 0.5468 & 0.5022 & 0.5438 & 0.5370 & 0.5470 & 0.5847 & 0.4979 & 0.6171 & 0.5910 & 0.5252 & 0.5483 \\
PLIP \cite{PLIP}         & 0.4637 & 0.5818 & 0.4454 & 0.4809 & 0.5455 & 0.5078 & 0.5384 & \underline{0.6427} & 0.5252 & \textbf{0.7298} & 0.5820 & 0.5676 & 0.5509 \\
CONCH \cite{CONCH}        & 0.5349 & \underline{0.6380} & 0.5638 & 0.5104 & 0.6384 & 0.5893 & \underline{0.6647} & 0.4935 & 0.5862 & 0.5608 & 0.6101 & 0.5680 & 0.5798 \\
Prov-GigaPath \cite{Prov-GigaPath} & 0.5633 & 0.6235 & 0.5544 & 0.5615 & 0.6184 & 0.6249 & 0.6403 & 0.6282 & 0.5092 & 0.7072 & 0.5943 & 0.6252 & 0.6042 \\
PathGen \cite{Pathgen} & 0.5184 & 0.4904 & 0.4750 & 0.4949 & 0.5807 & 0.5695 & 0.5708 & 0.4825 & 0.5197 & 0.4731 & 0.5099 & 0.4898 & 0.5146 \\
PathAlign \cite{Pathalign}   & 0.5875 & 0.6073 & \textbf{0.6865} & 0.6096 & 0.6276 & 0.6534 & \textbf{0.6693} & 0.6143 & 0.5871 & \underline{0.7208} & 0.6301 & \underline{0.6427} & 0.6364 \\
\midrule
LLaVA-Med \cite{Llava-med} & 0.5755 & 0.5770 & 0.5598 & 0.5690 & 0.5802 & 0.5748 & 0.5657 & 0.5711 & 0.5563 & 0.5875 & 0.5678 & 0.5694 & 0.5712 \\
MedDr \cite{Meddr} & 0.5433 & 0.5947 & 0.5245 & 0.5371 & 0.5924 & 0.6075 & 0.6096 & 0.6349 & 0.5817 & 0.7131 & 0.6240 & 0.5927 & 0.5963 \\
Quilt-LLaVA \cite{Quilt-llava} & 0.6122 & 0.6107 & 0.5981 & 0.5945 & 0.6077 & 0.6018 & 0.5962 & 0.6079 & 0.5896 & 0.6143 & 0.6002 & 0.6019 & 0.6029 \\
PathGen-LLava \cite{Pathgen} & 0.5860 & \textbf{0.6484} & 0.6252 & 0.6351 & 0.6427 & 0.6380 & 0.5482 & 0.6332 & 0.6247 & 0.6466 & 0.5699 & 0.6334 & 0.6193 \\
CPath-Omni \cite{Cpath-omni} & \underline{0.6423} & 0.6301 & 0.6355 & \underline{0.6430} & \textbf{0.6623} & \underline{0.6671} & 0.6520 & 0.6418 & \underline{0.6502} & 0.6389 & \underline{0.6309} & 0.6151 & \underline{0.6424} \\

\midrule

PathFLIP (w/o Region Q-Former) & 0.5152 & 0.6043 & 0.5432 & 0.5492  & 0.5651 & 0.5443 & 0.5034 & 0.5641 & 0.5034 & 0.5736 & 0.5066 & 0.5976 & 0.5475 \\

PathFLIP (w/o Slide Q-Former) & 0.5863 & 0.5450 & 0.5660 & 0.5467  & 0.5643 & 0.5486 & 0.5357 & 0.5698 & 0.5691 & 0.5721 & 0.5927 & 0.5279 & 0.5604 \\

PathFLIP (w/o T-R Atten.) & 0.5367 & 0.5921 & 0.5541 & 0.5025  & 0.5132 & 0.5226 & 0.5585 & 0.5229 & 0.5595 & 0.5898 & 0.5770 & 0.5036 & 0.5444 \\

\rowcolor[HTML]{EFEFEF} 
PathFLIP (Ours)
& \textbf{0.6575} & 0.6339 & \underline{0.6796}
& \textbf{0.6720} & \underline{0.6585} & \textbf{0.6713}
& 0.6452 & \textbf{0.6867} & \textbf{0.6646}
& 0.6209 & \textbf{0.6423} & \textbf{0.7287} & \textbf{0.6634} \\
\bottomrule
\end{tabular}%
}

\caption{Comparison of models on gene mutation prediction tasks across multiple cancer types in the CPTAC dataset. {Bold} indicates the best performance and {underline} indicates the second best performance. ``T-R Atten.'' refers to the Text-Region Attention layer.}
\label{tab:zero_shot}

\end{table*}

\section{Experiments}
\subsection{Experimental Setup}
\subsubsection{Datasets}

We use the SlideInstruction dataset \cite{Slidechat} for pretraining, which includes WSIs from the public TCGA database \cite{TCGA}, comprising 4,915 WSI-caption pairs from 4,028 patients. Each WSI is associated with a long and detailed, clinically meaningful caption.
To validate the PathFLIP performance for various tasks, we use the SlideBench dataset, which includes the remaining test set from TCGA and external datasets from BCNB \cite{BCNB}.
For zero-shot classification, we additionally use CPTAC \cite{CPTAC}, which contains genetic mutation information. 
We also employ the large-scale Quilt-1M \cite{Quilt1m} as an external benchmark for the retrieval task.

\subsubsection{Comparison Methods}
We compare PathFLIP with state-of-the-art methods for classification and retrieval tasks. These methods can be categorized into two groups.
The first group includes models without LLMs, which are typically used as foundation models (FMs) to extract task-agnostic features. Representative examples include CONCH \cite{CONCH}, PLIP \cite{PLIP}, and Prov-GigaPath \cite{Prov-GigaPath}.
The second group consists of VLMs designed for pathology and enhanced by LLMs. These models are often capable of following natural language instructions, enabling more flexible downstream applications. Representative methods include Quilt-LLaVA \cite{Quilt-llava}, LLaVA-Med \cite{Llava-med}, and PathGen-LLava \cite{Pathgen}.
Notably, these LLM-based methods are primarily developed for patch-level pathology images. In particular, Cpath-Omni \cite{Cpath-omni} and SlideChat \cite{Slidechat} incorporate dedicated slide encoders to generate a global slide-level representation as inputs to the LLMs.

\subsubsection{Implementation Details}
We implement PathFLIP in PyTorch 2.7.0 and conduct experiments on 8 NVIDIA RTX 4090 GPUs. Each WSI is partitioned into non-overlapping $4096 \times 4096$ regions at 20$\times$ magnification and further cropped into $256 \times 256$ patches for feature extraction using the pretrained CONCH \cite{CONCH}. We set the Q-Former query token count to $N_q = 8$ and sample $K = 8$ subcaptions for regional alignment. The temperature hyperparameter is set to $\tau = 0.1$, while the learnable parameter is initialized as $\eta = \log(1 / 0.07)$. The language module uses the lightweight Qwen3-0.6B \cite{Qwen3}, fine-tuned via LoRA \cite{Lora}.

\subsection{Comparisons with Previous Studies}

\begin{table*}
\centering

\resizebox{\textwidth}{!}{%
\begin{tabular}{l ccc ccc ccc ccc}
\toprule
\multicolumn{1}{c}{} & \multicolumn{6}{c}{SlideBench} & \multicolumn{6}{c}{Quilt} \\
\cmidrule(lr){2-7} \cmidrule(lr){8-13}
Model & \multicolumn{3}{c}{Image-Text Retrieval (ITR)} & \multicolumn{3}{c}{Text-Image Retrieval (TIR)} & \multicolumn{3}{c}{Image-Text Retrieval (ITR)} & \multicolumn{3}{c}{Text-Image Retrieval (TIR)} \\
 & R@1 & R@5 & R@10 & R@1 & R@5 & R@10 & R@1 & R@5 & R@10 & R@1 & R@5 & R@10 \\
\midrule
MI-Zero \cite{MI_Zero} & 0.0468 & 0.2120 & 0.2723 & 0.0491 & 0.2235 & 0.3725 & 0.1583 & 0.3316 & 0.4711 & 0.1516 & 0.3543 & 0.5036 \\

PLIP \cite{PLIP} 
&0.0468&0.2163&0.3725&0.0336&0.2053&0.3341& 0.1296 &0.2813 & 0.4208& 0.1152 &0.2751 &0.4092 \\

CONCH \cite{CONCH} &0.0360&0.2276&0.3798&{0.0576}&{0.2788}&{0.4399}& 0.2411 &0.4612& 0.6295& 0.2377 &0.4174 & \underline{0.6351} \\

Prov-GigaPath \cite{Prov-GigaPath} & 0.0468 & 0.2259 & {0.3822} & 0.0528 & 0.2235 & 0.3725 & 0.0123 & 0.1516 & 0.3070 & 0.0253 & 0.1482 & 0.3184 \\

PathGen \cite{Pathgen} &0.0513 &0.2165 &0.3629 &0.0535 &0.2139 &0.3581 & \textbf{0.3246} &\textbf{0.4955} & 0.6206 & \textbf{0.2688} &\textbf{0.4645} &0.6043 \\

PathAlign \cite{Pathalign} &0.0336 &{0.2299} &0.3341 &0.0432 &0.2139 &0.3629 & 0.1813 &0.3906 & 0.4169 & 0.1533 &0.3910 &0.5047 \\
\midrule

PathFLIP (w/o Region Q-Former) & 0.0581 & 0.2103 & 0.3613 & 0.0481 & 0.2099 & 0.3760 & 0.1795 & 0.3627 & 0.5654 & 0.1829 & 0.3567 & 0.5611 \\

PathFLIP (w/o Slide Q-Former) & \underline{0.1127} & \underline{0.3639} & \underline{0.5461} & 0.0811 & \underline{0.3357} & \underline{0.5504} & 0.1757 & 0.3333 & 0.6165 & 0.1719 & 0.3253 & 0.6293 \\

PathFLIP (w/o T-R Atten.) & 0.0781 & 0.2810 & 0.4823 & \underline{0.0861} & 0.3249 & 0.4979 & 0.1774 & 0.3278 & \underline{0.6301} & 0.1668 & 0.3257 & 0.6284 \\

\rowcolor[HTML]{EFEFEF}
PathFLIP (Ours) & \textbf{0.1513} & \textbf{0.3869} & \textbf{0.5704} & \textbf{0.1392} & \textbf{0.4380} & \textbf{0.5903} & \underline{0.2710} & \underline{0.4708} & \textbf{0.6611} & \underline{0.2646} & \underline{0.4291} & \textbf{0.6407} \\

\bottomrule
\end{tabular}%
}

\caption{Retrieval performance on the SlideBench and Quilt datasets for Image-Text Retrieval (ITR) and Text-Image Retrieval (TIR) tasks. The evaluation metric is Recall@K. {Bold} indicates the best performance in each column, and {underline} denotes the second best. All evaluations are conducted with a batch size of 64 from the dataset.}
\label{tab:retrieval_double}

\end{table*}

\begin{table}[h]

\centering
\setlength\tabcolsep{1.5pt}
\resizebox{\linewidth}{!}{
\begin{tabular}{lcccccccc}
\toprule
Methods  &  BLEU-1 & BLEU-2 & BLEU-3 & Rouge-L & \makecell[c]{Qwen.Em.\\Similarity} \\ \hline
LLava-Med$^\dag$ \cite{Llava-med} & 0.18 & 0.12& 0.06 & 0.12 & 0.58\\
MedDr$^\dag$ \cite{Meddr} & 0.31& 0.16&0.06& 0.07 & 0.68 \\
Quilt-LLava$^\dag$ \cite{Quilt-llava} & 0.23 &	0.09 &	0.04 &	0.16 & 0.56\\
PathGen-LLava$^\dag$ \cite{Pathgen} &0.35 &0.17 &0.09 &0.16 & 0.65\\
CPath-Omni \cite{Cpath-omni} &0.33 & 0.19 & 0.10 & 0.17 & 0.70\\
SlideChat \cite{Slidechat} & \underline{0.37}& \underline{0.21} &	\underline{0.12} &	0.24& \underline{0.71}\\
\hline
PathFLIP (w/o Region Q-Former) & 0.31 & 0.19 & \underline{0.12} & \underline{0.33} & 0.64 \\

PathFLIP (w/o Slide Q-Former) & 0.21 & 0.13 & 0.08 & 0.32 & 0.67 \\

PathFLIP (w/o T-R Atten.) & 0.32 & 0.19 & \underline{0.12} & \textbf{0.34} & 0.64 \\

PathFLIP (w/o finetune Qwen3)  & 0.25 & 0.14 & 0.09 & \underline{0.33} & 0.64 \\

\rowcolor[HTML]{EFEFEF} 
PathFLIP (w/ finetune Qwen3) & \textbf{0.38} & \textbf{0.23} & \textbf{0.13} & \textbf{0.34} & \textbf{0.76} \\
\bottomrule
\end{tabular}
}

\caption{WSI caption performance across different methods on SlideBench. Due to the inability to process the entire WSI simultaneously, methods marked with $^\dag$ use 30 randomly sampled patches from each slide as input. {Bold} indicates the best and {underline} denotes the second best.}
\label{tab:caption}

\end{table}

\subsubsection{Zero-shot Classification} 
We evaluate the generalisation capability of PathFLIP through zero-shot biomarker classification tasks on the CPTAC dataset, formulated as prompt-based tasks. Each task is formulated as a prompt-based classification problem, such as: \textit{“a histopathology image with a PIK3CA gene mutation.”}
The first group of baselines excludes LLMs, using averaged patch embeddings for slide-level representations and classifying based on similarity with text prompts.
The second group integrates LLMs but cannot directly handle gigapixel WSIs, so 30 patches are randomly sampled per slide and used with prompts for biomarker prediction.
As shown in Table~\ref{tab:zero_shot}, PathFLIP achieves the highest average score 0.6634, outperforming CPath-Omni and PathAlign by 2.1\% and 2.7\%, respectively, while using significantly fewer model parameters and less training data.
Further, when incorporating the Slide Q-Former, Region Q-Former, and text-region attention modules, the model achieves performance improvements of 11.59\%, 10.3\%, and 11.9\%, respectively, demonstrating the effectiveness of fine-grained alignment.
These results highlight the discriminative power of the slide-level representations at fine-grained and global levels.

\subsubsection{Zero-shot Retrieval}
We evaluate the multimodal alignment capability through the slide-text retrieval tasks on the SlideBench dataset and a region-level Quilt dataset, as summarised in Table~\ref{tab:retrieval_double}.
On SlideBench, our slide-level embeddings significantly outperform the mean patch embeddings of FMs, achieving an average improvement of 15.50\% on the ITR task and 13.04\% on the TIR task over CONCH.
We conduct region-level image-text retrieval tasks on the Quilt dataset to further validate the model's generative retrieval capability. PathFLIP achieves SOTA performance with 0.6611 and 0.6407 R@10 scores, respectively.
Compared to PathGen, which achieves R@1 scores of 0.3246 and 0.2688, our method ranks second on other evaluation metrics while being trained on a significantly smaller dataset, demonstrating the strong potential of our proposed PathFLIP.
The Region Q-Former module is crucial in retrieval performance, boosting average recall by 12.90\% for ITR task and 12.79\% for TIR task.
Among all comparison methods, PathFLIP shows superior capability in capturing fine-grained visual-language correlations, highlighting its effectiveness and efficiency, even with limited data resources.

\begin{table}[h]

\centering
\setlength\tabcolsep{2pt}
\resizebox{\linewidth}{!}{
\begin{tabular}{lccccc}
\toprule
\multirow{2}{*}{Methods} & \multicolumn{3}{c}{SlideBench}  & \multirow{2}{*}{\makecell[c]{BCNB\\VQA}} & \multirow{2}{*}{Average} \\
\cmidrule(lr){2-4}
& Microscopy & Diagnosis & Clinical \\
\midrule
LLava-Med$^\dag$ \cite{Llava-med}& 0.4734 & 0.3278 & 0.4796 & 0.3010 & 0.3955 \\
MedDr$^\dag$ \cite{Meddr}& 0.7330 & 0.5778 & 0.7425 & 0.3367 & 0.5975 \\
Quilt-LLava$^\dag$ \cite{Quilt-llava}& 0.5776 & 0.3596 & 0.5307 & 0.3219 & 0.4475 \\
PathGen-LLava$^\dag$ \cite{Pathgen} & 0.6867 & 0.4969 & 0.6336 & 0.4557 & 0.5682 \\
CPath-Omni \cite{Cpath-omni}& 0.6370 & 0.5241 & 0.5926 & 0.4567 & 0.5526 \\
SlideChat \cite{Slidechat} & \textbf{0.8764} & \underline{0.7327} & \underline{0.8426} & \underline{0.5414} & \underline{0.7483} \\
\midrule
\rowcolor[HTML]{EFEFEF}
Ours (w/ finetune Qwen3) & \underline{0.8611} & \textbf{0.7628} & \textbf{0.8941} & \textbf{0.5865} & \textbf{0.7761} \\
\bottomrule
\end{tabular}
}

\caption{VQA performance of different methods on the SlideBench and BCNB. 
Methods marked with $^\dag$ use 30 randomly sampled patches from each slide.
{Bold} indicates the best performance and {underline} denotes the second best.
}
\label{tab:vqa}

\end{table}

\subsubsection{WSI Caption and Visual Question Answering}
As shown in Table~\ref{tab:caption} and Table~\ref{tab:vqa}, we evaluate caption generation performance using BLEU, ROUGE-L, and Qwen3-Embedding-0.6B \cite{Qwen3-Em} for similarity scoring.
We observe that Quilt-LLava and PathGen-LLava, which rely on patch-level inputs, perform worse than models capable of holistically processing the entire slide.
Notably, our PathFLIP significantly outperforms SlideChat, achieving a 10\% improvement in Rouge-L and a 5\% gain in Qwen embedding similarity in the captioning task. 
For the VQA task, PathFLIP also outperforms SlideChat with average improvements of 2.78\%. Specifically, on the external BCNB dataset, PathFLIP achieves a strong score of 0.58.
These results demonstrate that our model can efficiently generate descriptive captions for WSIs and effectively interpret complex instructions.
Furthermore, when applying instruction fine-tuning on a lightweight model like Qwen3-0.6B, we observe a 12\% improvement in similarity scores for captioning.
These findings underscore the model's strong multimodal understanding and instruction-following capabilities for gigapixel-scale WSIs. They also highlight the practical potential of our pretraining approach when integrated with lightweight LLMs in real-world clinical applications.
Additional experimental results are in supplementary materials.

\section{Conclusion}
We propose PathFLIP, a fine-grained language-image pretraining framework for general-purpose pathological image analysis. By aligning region-level visual features with subcaption semantics via contrastive learning, PathFLIP achieves accurate and interpretable grounding without manual annotations. Integrated with large language models, PathFLIP supports diverse multimodal tasks such as WSI captioning and VQA, and consistently outperforms existing pathology VLMs on multiple benchmarks. Beyond technical advancements, PathFLIP facilitates interpretable AI-assisted diagnosis and clinician-AI communication, paving the way for autonomous AI agents in digital pathology to enhance clinical decision-making and healthcare.

\section{Acknowledgments}
This work was supported in part by the National Natural Science Foundation of China under 62031023 \& 62331011; and in part by the Shenzhen Science and Technology Project under GXWD20220818170353009.

\bibliography{aaai2026}

\clearpage
\section{Supplementary Material}

\subsection{Model Details}
\begin{figure}[t]
\centering
\includegraphics[width=0.80\linewidth]{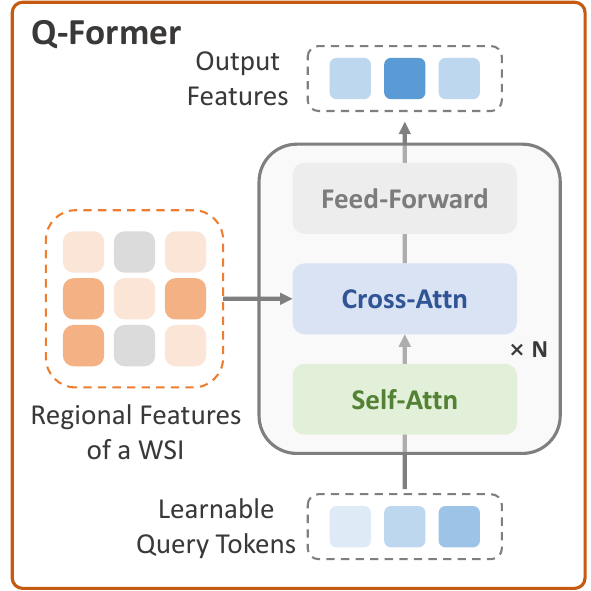}
\caption{
\textbf{Overview of the Region Q-former module}, which consists of a self-attention layer, a cross-attention layer, and a feed-forward layer.
}
\label{Q-former}
\end{figure}

Figure~\ref{Q-former} illustrates the architecture of the Q-Former module. We introduce both the Region Q-former and the Slide Q-Former, which share a similar structure and enable attention-based feature aggregation. These modules play a key role in facilitating effective feature interaction at both the region and slide levels.
The Q-Former modules take a set of learnable queries as input to perform feature compression. Specifically, the Region Q-former operates on individual region embeddings (as shown in Figure~\ref{Q-former}), while the Slide Q-Former processes a set of patch embeddings. Through this process, we obtain compressed representations that reduce feature dimensionality while promoting more efficient interaction between features.

\subsection{Datasets Details}
\subsubsection{SlideInstruct and SlideBench}

\begin{figure*}[!t]
\centering
\includegraphics[width=\linewidth]{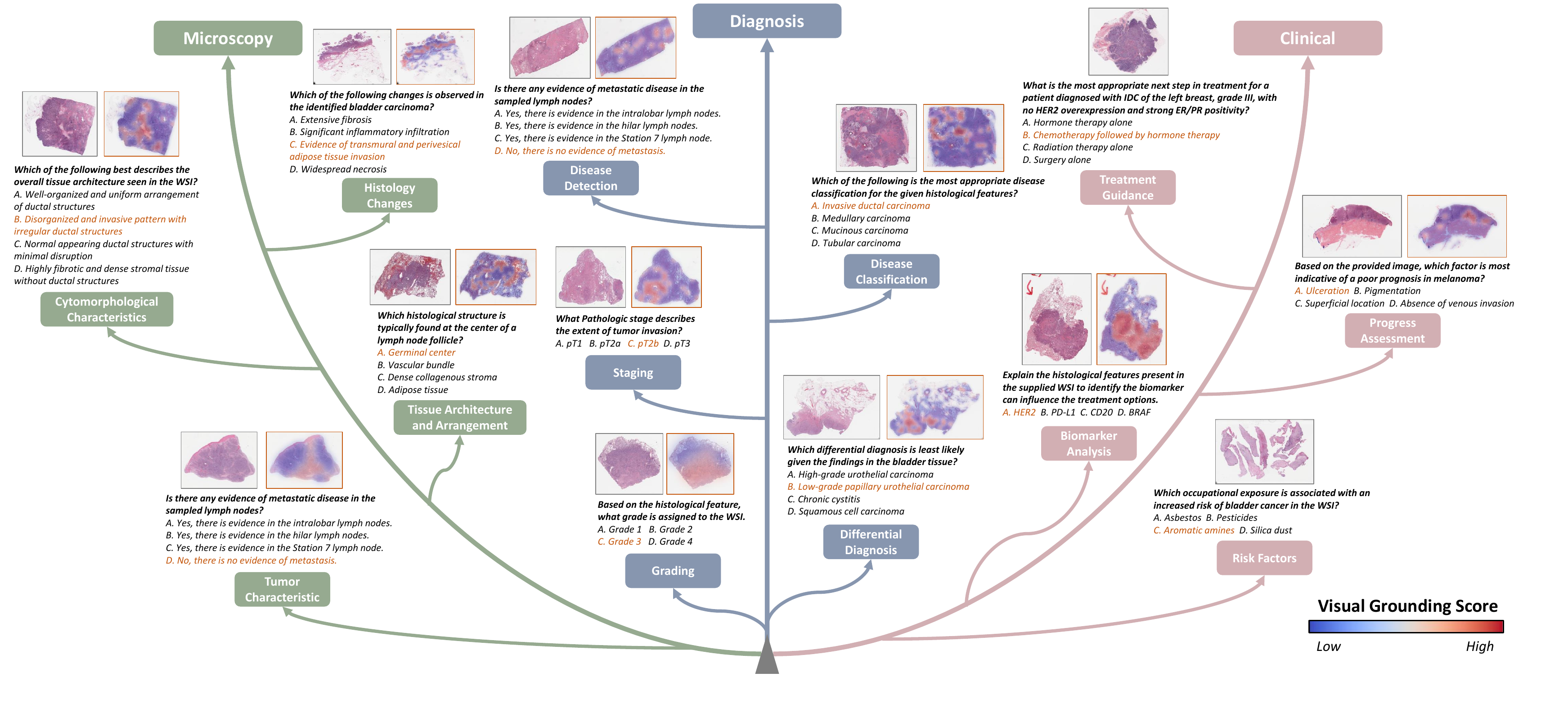}
\caption{
\textbf{PathFLIP handles visual question answering and visual grounding simultaneously in VQA tasks}. High-attention areas are highlighted in red on the heatmaps. The examples cover three major VQA categories in pathology: \textit{Microscopy}, \textit{Diagnosis}, and \textit{Clinical}.
}
\label{vqa1}
\end{figure*}

We use the publicly available SlideInstruct dataset for training and the SlideBench dataset as the validation benchmark.
SlideInstruct consists of 4,915 WSI-captions pairs from 4,028 patients, with each WSI sourced from The Cancer Genome Atlas (TCGA) database. This dataset is composed of two parts: WSI-Caption and WSI-Instruction data. The WSI-Instruction data includes textual descriptions and open-set Visual Question Answering (VQA) questions designed to evaluate model understanding and reasoning.
SlideBench consists of two subsets: SlideBench-TCGA and SlideBench-BCNB.
To evaluate model performance, we use SlideBench-TCGA, which contains 734 WSI-caption pairs paired with the corresponding closed-set VQA samples for testing. The VQA questions in SlideBench-TCGA are categorized into three types: Microscopy, Diagnosis, and Clinical. Some samples and the results generated by PathFLIP are shown in Figure~\ref{vqa1}
Additionally, the SlideBench-BCNB dataset includes 1,058 WSIs with slide-level VQA pairs, which are used for zero-shot VQA evaluation.
TCGA WSIs can be accessed from: \url{https://proteomics.cancer.gov/data-portal},
BCNB WSIs are available at: \url{https://bupt-ai-cz.github.io/BCNB/},
and SlideBench and SlideInstruct datasets are available at: \url{https://huggingface.co/datasets/General-Medical-AI/SlideChat}.

\subsubsection{CPTAC}
We collected a total of 444 primary cases from the CPTAC dataset, which includes 120 cases of Breast Invasive Carcinoma (BRCA), 110 cases of Colon Adenocarcinoma (COAD), 106 cases of Lung Adenocarcinoma (LUAD), and 108 cases of Lung Squamous Cell Carcinoma (LUSC). For each case, we also retrieved the corresponding molecular status, including the following mutations:
\textit{PIK3CA}, \textit{MAP3K1}, and \textit{GATA3} for BRCA,
\textit{KRAS}, \textit{PIK3CA}, and \textit{TP53} for COAD,
\textit{TP53}, \textit{STK11}, and \textit{KRAS} for LUAD, and
\textit{TP53}, \textit{PIK3R1}, and \textit{KEAP1} for LUSC.
The CPTAC dataset can be accessed at: \url{https://proteomic.datacommons.cancer.gov/pdc/}.

\subsubsection{Quilt-1M}

Quilt-1M is a region-level dataset consisting of image-text pairs with a resolution of 512$\times$512, designed for retrieval tasks. It contains valuable educational histopathology content curated from expert clinicians, with many examples sourced from YouTube videos.
The dataset is available for download at: \url{https://huggingface.co/datasets/wisdomik/QUILT-LLaVA-Instruct-107K}.

\subsection{Experiment Details}
\subsubsection{Zero-shot Classification Prompts}
To evaluate the performance of zero-shot classification in gene mutation status, we construct a set of prompt templates that serve as textual queries. These templates are shared across different gene mutation classification tasks. For each sample, a random subset of prompts is selected and ensembled in the text embedding space to improve robustness.

For cases where a gene is mutated, the following prompts are used:
\begin{itemize}
  \item \texttt{A histopathology image with a \textit{CLASSNAME} gene mutation.}
  \item \texttt{This whole slide image shows signs of a \textit{CLASSNAME} mutation.}
  \item \texttt{This is a histopathology slide from a patient with a \textit{CLASSNAME} mutation.}
  \item \texttt{This tissue sample is positive for a \textit{CLASSNAME} mutation.}
  \item \texttt{The \textit{CLASSNAME} gene is mutated in this sample.}
\end{itemize}

For cases where the gene is not mutated, the following prompts are used:
\begin{itemize}
  \item \texttt{A histopathology image without a \textit{CLASSNAME} gene mutation.}
  \item \texttt{This whole slide image shows no evidence of a \textit{CLASSNAME} mutation.}
  \item \texttt{This is a histopathology slide from a patient without a \textit{CLASSNAME} mutation.}
  \item \texttt{This tissue sample is negative for a \textit{CLASSNAME} mutation.}
  \item \texttt{The \textit{CLASSNAME} gene is not mutated in this sample.}
\end{itemize}

Here, \texttt{\textit{CLASSNAME}} is replaced by the target gene name in each classification task. This zero-shot setup highlights the model's ability to understand the multimodal relationships between histopathology images and textual descriptions.

\subsubsection{Instruct Fine-tuning Prompts}
We combine PathFLIP with Qwen3-0.6B for further instruction fine-tuning, adopting an instruction-based prompting strategy on the SlideInstruct dataset. Each input consists of a system prompt and a user query. The system prompt is fixed as follows:

\texttt{\textbf{[System Prompt:]} You are an AI assistant proficient in digital pathology.}

The \texttt{\textbf{[User Prompt]}} is selected randomly from a predefined pool of instruction templates, such as:

\begin{table*}[!t]
\centering
\caption{Few-shot (ratio = 0.05) comparison of models on gene mutation prediction tasks across multiple cancer types in the CPTAC dataset. \textbf{Bold} indicates the best performance and \underline{underline} indicates the second best performance. `T-R Atten.'' refers to the Text-Region Attention layer.}
\label{tab:zero_shot1}
\resizebox{\textwidth}{!}{%
\begin{tabular}{l lll lll lll lll c}
\toprule
 AUC[\%] & \multicolumn{3}{c}{BRCA} & \multicolumn{3}{c}{COAD} & \multicolumn{3}{c}{LUAD} & \multicolumn{3}{c}{LSCC} & \multirow{2}{*}{Average} \\
\cmidrule(lr){2-4}
\cmidrule(lr){5-7}
\cmidrule(lr){8-10}
\cmidrule(lr){11-13}
 Model & \textit{PIK3CA} & \textit{MAP3K1} & \textit{GATA3} & \textit{KRAS} & \textit{PIK3CA} & \textit{TP53} & \textit{TP53} & \textit{STK11} & \textit{KRAS} & \textit{TP53} & \textit{PIK3R1} & \textit{KEAP1} & \\
\midrule
MI-Zero & 0.4461 & 0.5719& 0.5324 & 0.5583 & 0.5585 & 0.5720 & 0.6358&0.6339 & 0.4944 & 0.5232 & 0.5157 & 0.5530 & 0.5496 \\
PLIP & 0.5717 & \textbf{0.7143}&0.4831 & 0.5874 & 0.5979 & 0.5381 & 0.5900 & 0.5952 & 0.5477 & 0.6275 & 0.4567 & 0.5845 & 0.5745 \\
CONCH & 0.5835 & 0.5690& 0.6130 & 0.6072 & 0.5756 & \underline{0.6323} & 0.6341 & 0.5736 & \underline{0.6270} & 0.6666 & 0.5445 & 0.5248 & 0.5959 \\
Prov-GigaPath & 0.5510 & 0.5996 & 0.6544 & 0.5423 & \underline{0.6347} & 0.6101 & 0.5427 & 0.6425 & 0.5880 & 0.5354 & 0.5468 & 0.6402 & 0.5906 \\
PathGen & \textbf{0.8190} & 0.5384 & 0.5851 & 0.5757 & 0.5463 & 0.5723 & 0.5677 & \textbf{0.6517} & 0.5295 & \textbf{0.7121} & 0.5441 & 0.5133 & 0.5963 \\
PathAlign & 0.5919 & 0.6086 & \underline{0.6883} & \underline{0.6346} & 0.6153 & 0.6259 & \textbf{0.6689} & 0.6299 & 0.6185 & \underline{0.7021} & \textbf{0.6414} & 0.6511 & \underline{0.6397} \\
\midrule
LLaVA-Med & 0.5815 & 0.6093& 0.6458 & 0.4869 & 0.5866& 0.5457 & 0.4962&0.4967& 0.6011 & 0.5069 & 0.5288 & 0.5035& 0.5491 \\
MedDr & 0.5344 & 0.5166 & 0.5172 &0.5198 & 0.5304 & 0.5482 & 0.5493 & 0.5310 & 0.5487 &0.5862 & 0.5482& 0.6750 & 0.5504 \\
Quilt-LLaVA & 0.5528 & 0.6349& 0.5804 & 0.6086 &0.5962 & 0.5137 &0.5428 &0.6034 &0.5047 &0.5129 &  0.5283& 0.5714& 0.5625  \\
CPath-Omni & 0.5344 &0.5530& 0.5625 & 0.5518 & 0.5647 & 0.5432 &0.5586 & 0.5326 &0.5292 & 0.5882 & 0.5537 & 0.6739 & 0.5622 \\
\midrule
PathFLIP (w/o Region Q-Former) & 0.6295 & 0.4753& 0.5787 &0.5105& 0.5275& 0.5650 & 0.5293 & 0.6276& 0.5110& 0.5393 & 0.4637 & 0.6363 & 0.5495 \\

PathFLIP (w/o Slide Q-Former) & 0.5393& 0.4752& 0.6789 & 0.5606& 0.5621& 0.5553& 0.5030 & 0.5758 & 0.5436&0.5121 & 0.5239 & 0.6363 & 0.5555 \\

PathFLIP (w/o T-R Atten.) & 0.5642& 0.5554& 0.5907 &0.5411 & 0.6311& 0.5857& 0.5119 & 0.5164 & 0.5710 & 0.5242 & 0.5046 & \textbf{0.8000} & 0.5747 \\
\rowcolor[HTML]{EFEFEF} 
PathFLIP (Ours)
& \underline{0.6389} & \underline{0.6544} & \textbf{0.6954} & \textbf{0.6426}& \textbf{0.6441} & \textbf{0.6484} & \underline{0.6366} & \underline{0.6455} & \textbf{0.6321} & 0.6242 &\underline{0.6041} & \underline{0.7380} & \textbf{0.6504} \\
\bottomrule
\end{tabular}%
}
\end{table*}

\begin{table*}[!t]
\centering
\caption{Few-shot (ratio = 0.1) comparison of models on gene mutation prediction tasks across multiple cancer types in the CPTAC dataset. \textbf{Bold} indicates the best performance and \underline{underline} indicates the second best performance. `T-R Atten.' refers to the Text-Region Attention layer.}
\label{tab:zero_shot2}
\resizebox{\textwidth}{!}{%
\begin{tabular}{l lll lll lll lll c}
\toprule
 AUC[\%] & \multicolumn{3}{c}{BRCA} & \multicolumn{3}{c}{COAD} & \multicolumn{3}{c}{LUAD} & \multicolumn{3}{c}{LSCC} & \multirow{2}{*}{Average} \\
\cmidrule(lr){2-4}
\cmidrule(lr){5-7}
\cmidrule(lr){8-10}
\cmidrule(lr){11-13}
 Model & \textit{PIK3CA} & \textit{MAP3K1} & \textit{GATA3} & \textit{KRAS} & \textit{PIK3CA} & \textit{TP53} & \textit{TP53} & \textit{STK11} & \textit{KRAS} & \textit{TP53} & \textit{PIK3R1} & \textit{KEAP1} & \\
\midrule
MI-Zero & 0.5451 & 0.5902& 0.6418 & 0.5969 & 0.6279 & \underline{0.6666} & 0.5372 & 0.5394 & 0.6063 & 0.5103 & 0.5333 & 0.6180 & 0.5844 \\
PLIP & 0.5647 & 0.7747& 0.6117 & 0.6380 & \textbf{0.6959} & 0.5793 & 0.6235 & 0.6804 & \underline{0.6541} & 0.5989 & 0.5434 & 0.6555 & 0.6350 \\
CONCH & 0.5876 & \textbf{0.7932} & \textbf{0.7512} & 0.5663 & 0.6277 & 0.6630 & 0.6028 & 0.7018 & 0.6177 & \textbf{0.8645} & 0.5887 & 0.5777 & \underline{0.6619}  \\
Prov-GigaPath & \underline{0.6496} & 0.6640 & 0.5285 & 0.5707 & 0.6439 & 0.6428 & 0.6068 & 0.6025 & 0.6055 & 0.5208 & 0.6711 & 0.6388 & 0.6121  \\
PathGen & 0.6077 & 0.7165& 0.6454 & 0.5088 & 0.6341 & 0.5716 & 0.5993 & 0.5773 & 0.6040 & 0.4843 & 0.5384 & 0.5487 & 0.5863 \\
PathAlign & 0.6085 & 0.6294& 0.6940 & \underline{0.6468} & 0.6369 & 0.6617 & \textbf{0.6747}& 0.6388 & 0.6259 & 0.7250 & 0.6447 & \underline{0.6633} & 0.6541 \\
\midrule
LLaVA-Med & 0.5861 & 0.6315& 0.6384 & 0.5770&0.5681 & 0.5625 &0.5515 & 0.5265 & 0.5679& 0.6941 &\textbf{0.7040}& 0.5306 & 0.5949 \\
MedDr & 0.5923 & 0.5679& 0.5974 & 0.5526 & 0.5745 &0.5828 & 0.5468 & 0.5526& 0.5330 & 0.6067& 0.5649 & 0.5664 & 0.5698  \\
Quilt-LLaVA & 0.5566 & 0.6438& 0.6105 &0.5478 & 0.5681 & 0.5645 & 0.5319 & 0.5529  & 0.6312& 0.5558 & 0.6121 & 0.6354 & 0.5842 \\
CPath-Omni & 0.5568& 0.5609& 0.5702 & 0.5789 & 0.5829&0.5289 &0.5234& 0.5682 & 0.6177 & 0.5563 &0.5585 &0.6254& 0.5690  \\
\midrule
PathFLIP (w/o Region Q-Former) & 0.5482 & 0.5039 & 0.5480 & 0.5312 & 0.5227 & 0.5604 & 0.5878 & \textbf{0.7155} & 0.5127 & 0.6276 & 0.6675 & 0.6034 & 0.5774 \\

PathFLIP (w/o Slide Q-Former) & 0.5902 &0.6087 & 0.5805 & 0.5787 & 0.5657 & 0.5866 & 0.5305 & 0.5949 & 0.5755 & 0.5103 & 0.5579 & 0.5308 & 0.5675 \\

PathFLIP (w/o T-R Atten.) & 0.5348 & 0.4686 & 0.5400 &0.5931 &0.5413 &0.5014 & 0.5046 & 0.6320 & 0.5819 & \underline{0.7500} & 0.5181 & 0.6368 & 0.5669  \\

\rowcolor[HTML]{EFEFEF} 
PathFLIP (Ours)
 & \textbf{0.6565} & \underline{0.7820}& \underline{0.7093} & \textbf{0.6549} & \underline{0.6625} & \textbf{0.6730} & \underline{0.6482} & \underline{0.7087} & \textbf{0.6622} & 0.6519 & \underline{0.6721} & \textbf{0.6791} & \textbf{0.6800} \\
\bottomrule
\end{tabular}%
}
\end{table*}

\begin{table*}[t]
\centering
\caption{Few-shot (ratio = 0.3) comparison of models on gene mutation prediction tasks across multiple cancer types in the CPTAC dataset. \textbf{Bold} indicates the best performance and \underline{underline} indicates the second best performance. `T-R Atten.'' refers to the Text-Region Attention layer.}
\label{tab:zero_shot3}
\resizebox{\textwidth}{!}{%
\begin{tabular}{l lll lll lll lll c}
\toprule
 AUC[\%] & \multicolumn{3}{c}{BRCA} & \multicolumn{3}{c}{COAD} & \multicolumn{3}{c}{LUAD} & \multicolumn{3}{c}{LSCC} & \multirow{2}{*}{Average} \\
\cmidrule(lr){2-4}
\cmidrule(lr){5-7}
\cmidrule(lr){8-10}
\cmidrule(lr){11-13}
 Model & \textit{PIK3CA} & \textit{MAP3K1} & \textit{GATA3} & \textit{KRAS} & \textit{PIK3CA} & \textit{TP53} & \textit{TP53} & \textit{STK11} & \textit{KRAS} & \textit{TP53} & \textit{PIK3R1} & \textit{KEAP1} & \\
\midrule
MI-Zero & 0.6219 &0.5834& 0.6450 & 0.5864 & 0.6190 & 0.6083 & 0.5597 & 0.5383 & 0.5894 & 0.5672 & 0.6576 & 0.6268 & 0.6003 \\
PLIP & 0.5844 & 0.5899 & 0.6010 & 0.6115 & \textbf{0.7507} & 0.5406 & 0.5942 & 0.5798 & 0.5628 & 0.6176 & 0.5617 & 0.5783 & 0.5977  \\
CONCH & 0.6048 & 0.5934& \textbf{0.7766} & 0.5444 & 0.5376 & 0.6475 & 0.6081 & \textbf{0.7386} & \textbf{0.6818} & 0.6294 & 0.6044 & 0.5988 & 0.6305  \\
Prov-GigaPath & 0.5612 & 0.6098& 0.5869 & \underline{0.6545} & 0.5967 & 0.5729 & 0.6157 & 0.7183 & 0.5948 & 0.5361 & 0.5852 & 0.6433 & 0.6063  \\
PathGen & 0.6231 & 0.6312& 0.6407 & 0.5556 & 0.5602 & 0.5421 & 0.5675 & 0.6046 & 0.5358 & 0.5375 & 0.5625 & 0.5798 & 0.5784 \\
PathAlign & 0.6178 & 0.6228 & 0.6833 & 0.6505 & 0.6779 & \textbf{0.6791} & \underline{0.6540} & 0.6679 & 0.6554 & \textbf{0.6895} & \underline{0.6647} & \underline{0.6731} & \underline{0.6613}  \\
\midrule
LLaVA-Med & 0.6319 & 0.6627& 0.6607 & 0.5868 & 0.5879 & 0.5850 & 0.5742 & 0.6071&0.6704 & 0.5937 & 0.5628& 0.5807 & 0.6087 \\
MedDr & 0.6072 & 0.6160 & 0.6237 & 0.5994 & 0.5978 & 0.5643 & 0.5473 & 0.6629 & 0.6363 & 0.5760 & 0.5594 & 0.5739 & 0.5970  \\
Quilt-LLaVA &\underline{0.6407} &\underline{0.7020}& 0.6494& 0.5606 & 0.6048 & 0.5684 & 0.5909 & 0.5562 &0.5799 & 0.5521 & 0.5764 & 0.5876 & 0.5974  \\
CPath-Omni &0.6190 & 0.6078& 0.6570 & 0.5588 & 0.6101 & 0.5647 & 0.5708 & 0.5868 & 0.5975& 0.5588 & 0.5881 & 0.6098 & 0.5941 \\
\midrule

PathFLIP (w/o Region Q-Former) & 0.5600 & 0.5615& 0.4891 & 0.5441 & 0.5335 & 0.5700 & 0.5823 &0.7089 & 0.5405& 0.5296 & 0.5662 & 0.5660 & 0.5626 \\

PathFLIP (w/o Slide Q-Former) & 0.5583 & 0.5770& 0.5208 &0.5792 & 0.6071 & 0.5516 &0.5291 & 0.5633 & 0.5631 & 0.5343 & 0.5938&0.5776 & 0.5646 \\

PathFLIP (w/o T-R Atten.) &0.5509& 0.5686& 0.6043 & 0.6502 & 0.6130 & 0.5561& 0.5141& 0.5680 &0.5417 &0.5441 & 0.5634 & 0.5697 & 0.5703  \\
 
\rowcolor[HTML]{EFEFEF} 
PathFLIP (Ours)
 & \textbf{0.6802} &\textbf{0.7681}& \underline{0.7423} & \textbf{0.6654} & \underline{0.6806} & \underline{0.6571}& \textbf{0.6635}& \underline{0.7212} & \underline{0.6756}& \underline{0.6682} & \textbf{0.6721} & \textbf{0.6844} & \textbf{0.6899} \\
\bottomrule
\end{tabular}%
}
\end{table*}

\begin{itemize}
  \item \texttt{Describe the key pathological features identified in the analysis of this whole slide image. <image>}
  \item \texttt{Provide a detailed pathological description of the tissue in this slide. <image>}
  \item \texttt{Summarize the diagnostic features observed in the whole slide image. <image>}
  \item \texttt{What abnormalities can be observed in this histopathology slide? <image>}
  \item \texttt{Explain the histological characteristics present in this sample. <image>}
  \item \texttt{Describe the cell morphology and tissue structure in this WSI. <image>}
  \item \texttt{Write a clinical interpretation of this histopathological image. <image>}
  \item \texttt{List key diagnostic indicators in this whole slide image. <image>}
  \item \texttt{What is the most likely diagnosis based on this tissue image? <image>}
   \item \texttt{Write a pathology-style impression paragraph based on the analysis of this pathology slide. <image>}
\end{itemize}

Here, \texttt{<image>} denotes the placeholder for the image input, which is internally replaced with image embeddings during training. This prompt design encourages the model to generate detailed and diverse captions or diagnostic descriptions conditioned on whole slide image content.

\subsubsection{Sample Sub-captions}
\begin{figure}[t]
\centering
\includegraphics[width=0.80\linewidth]{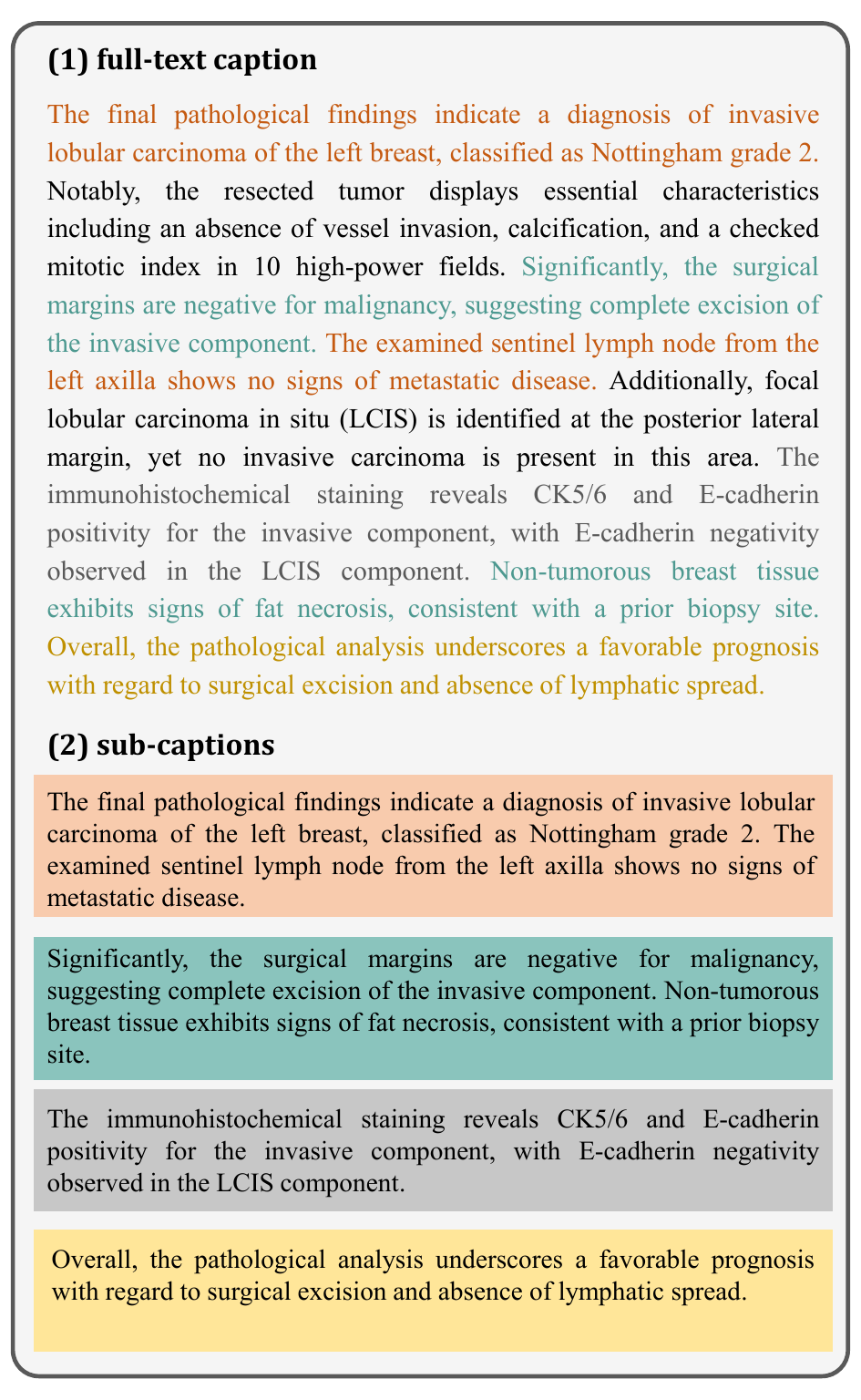}
\caption{
An example from sample \textbf{TCGA-A7-A3J1-01Z-00-DX1}, illustrating: (1) its full-text caption, and (2) the $K$ generated sub-captions. Here, $K=4$ is used as an example.
}
\label{sub-caption}
\end{figure}

As shown in Figure \ref{sub-caption}, each sub-caption contains 1–3 sentences. We first segment the full-text caption into individual sentences. Then, each sentence is selected with equal probability, and the selected sentences are merged into a sub-caption. In our experiments, we set $K=8$. At each iteration, the sub-captions vary, which serves as a form of data augmentation.

\subsubsection{Experimental Settings}
Experiments are conducted on four NVIDIA RTX 4090 GPUs with a batch size of 8 per device. We use the AdamW optimizer with weight decay of 0.05, and a learning rate scheduler combining 400-step linear warmup and cosine decay, with peak and minimal learning rates of \(1 \times 10^{-4}\) and \(5 \times 10^{-6}\), respectively. Training follows a 5-fold cross-validation setup, with early stopping based on the lowest validation loss.
The model is pre-trained for 100 epochs, followed by 10 epochs of instruction fine-tuning using Qwen3-0.6B. LoRA parameters are set to \(r = 8\), \(\alpha = 32\), and dropout = 0.1. The total computation time is 100 GPU hours for pre-training and 20 GPU hours for fine-tuning using BFloat16 mixed precision.

\begin{figure*}[!t]
\centering
\includegraphics[width=\linewidth]{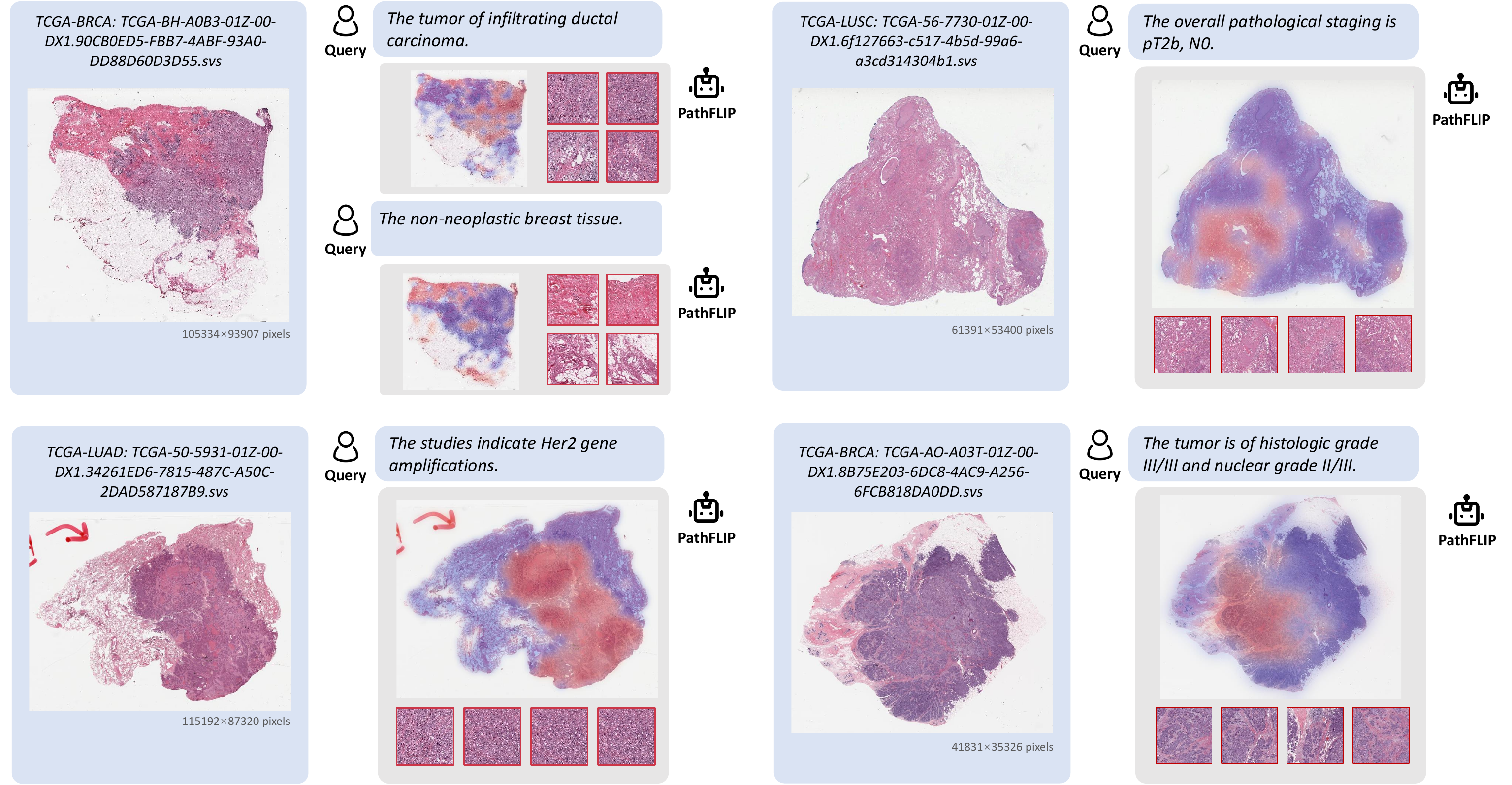}
\caption{
\textbf{Visual grounding results}. High-attention areas are highlighted in red in the heatmap, with red boxes marking the corresponding regions.
}
\label{grounding1}
\end{figure*}

\subsubsection{Evaluation Metrics}

For the Image-Text Retrieval (ITR) and Text-Image Retrieval (TIR) tasks, we use Recall@$K$ as the evaluation metric, which measures the percentage of relevant items successfully retrieved among the top $K$ candidates.
For classification tasks, we report the Area Under the Curve (AUC) as the main evaluation metric. In the VQA task, since all questions in the benchmark are in a closed-set format, we use accuracy (ACC) to assess performance.

In the caption generation task, we adopt BLEU and Rouge-L scores as standard evaluation metrics. Additionally, we introduce a novel evaluation approach using Qwen3-Embedding-0.6B, a language model that extracts semantic embeddings from both the ground-truth captions and the generated captions. The similarity between these embeddings is then computed as a score ranging from 0 to 1. Compared to BLEU and Rouge-L, which primarily focus on n-gram overlaps, this embedding-based metric better captures the semantic relevance between texts, providing a more meaningful assessment of caption quality.

\subsection{Additional Experimental Results}
\subsubsection{Few-shot Classification}
We conducted a series of few-shot classification experiments for the CPTAC gene mutation prediction task, which is framed as a binary classification problem. This task is characterized by significant class imbalance, with the ratio of positive to negative samples ranging from 1:6 to as high as 1:10 for most genes.
For each gene, we sampled training instances according to the specified few-shot ratio. To address the severe class imbalance, we ensured that at least one positive sample was included in the training set when the minority class had very few examples. 
Notably, due to the class imbalance, increasing the few-shot ratio does not always lead to higher AUC scores when the remaining samples are used for evaluation, which may exacerbate the class imbalance.

We performed experiments on four CPTAC gene mutation datasets. As shown in Tables~\ref{tab:zero_shot1}, \ref{tab:zero_shot2}, and \ref{tab:zero_shot3}, we evaluated model performance at few-shot ratios of 0.05, 0.1, and 0.3. Slide-level visual features were computed by averaging patch embeddings extracted from the vision encoder. For models incorporating a large language model (LLM), features were extracted from the vision tower.
A multilayer perceptron (MLP) classifier was trained with early stopping to select the optimal checkpoint. All models were trained under identical conditions, including the same optimizer and a fixed number of 50 training epochs.
The experimental results demonstrate that PathFLIP consistently outperforms baseline models in few-shot slide-level classification, achieving the best AUC performance of 0.6504, 0.6800, and 0.6899 for few-shot ratios of 0.05, 0.1, and 0.3, respectively. This highlights the model's strong performance across all few-shot settings.


\subsubsection{Visual Grounding}
We further explore the interpretability and visual grounding abilities of our model in the VQA task. As shown in Figure~\ref{vqa1}, our model not only delivers accurate answers to a wide range of VQA questions but also provides diagnostic reasoning to support its responses.
Our model addresses three major categories commonly studied in pathology-related VQA tasks: Microscopy, Diagnosis, and Clinical. For instance, in the Microscopy category, it handles questions about tumor characteristics, cytomorphological features, tissue architecture, and histologic changes.
Leveraging attention heatmaps, our model can highlight relevant regions within whole-slide images, offering spatial grounding for its answers.

Figure~\ref{grounding1} presents examples demonstrating the subcaption-level visual grounding capabilities of PathFLIP. It accurately localizes morphological features, such as ``infiltrating ductal carcinoma" and ``non-neoplastic breast tissue."
Moreover, the model exhibits a reasoning process that closely mirrors that of pathologists. For example, it can infer the underlying logic behind pathological staging decisions and provide detailed insights into histologic and nuclear grades.

\begin{figure*}[t]
\centering
\includegraphics[width=\linewidth]{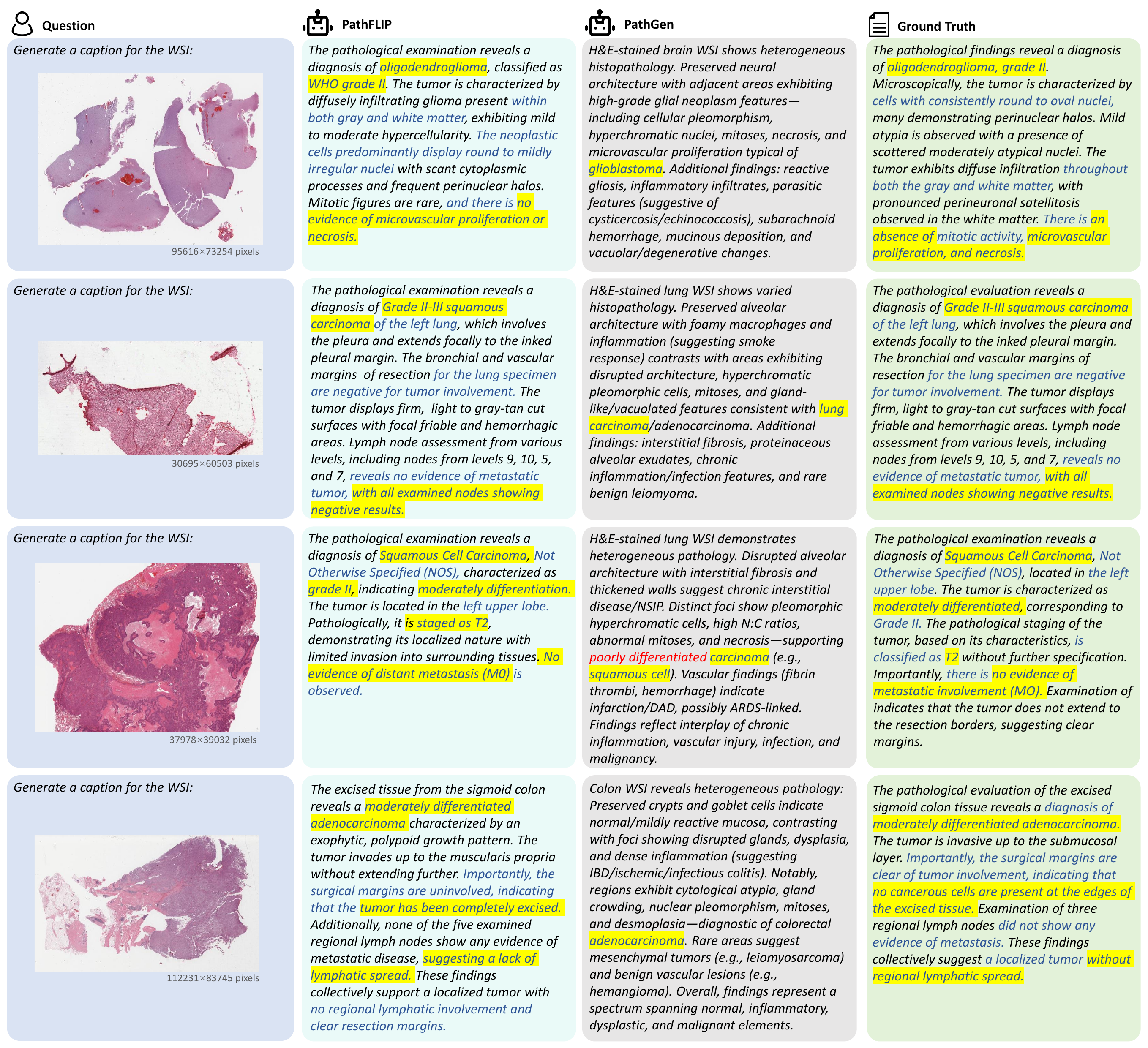}
\caption{\textbf{Caption generation comparison}. Blue indicates correct matches, red indicates incorrect or imprecise matches, and yellow backgrounds emphasize important information matches.}
\label{caption}
\end{figure*}

Currently, research lacks quantitative metrics for evaluating the accuracy of visual grounding and segmentation, primarily due to the absence of publicly available pathology datasets that include expert-annotated regions paired with corresponding textual explanations related to morphological or diagnostic features. This contrasts with the natural image domain, where such annotated datasets are more readily available.
In the future work, we aim to extend our method to facilitate the development of benchmark datasets addressing this gap, particularly those annotated with Regions of Interest (ROIs) and key diagnostic regions identified by pathologists. Our visual grounding module has the potential to significantly reduce the annotation burden, thereby enabling the creation of large-scale, expert-curated datasets and accelerating progress in computational pathology.

\begin{figure*}[!t]
\centering
\includegraphics[width=0.95\linewidth]{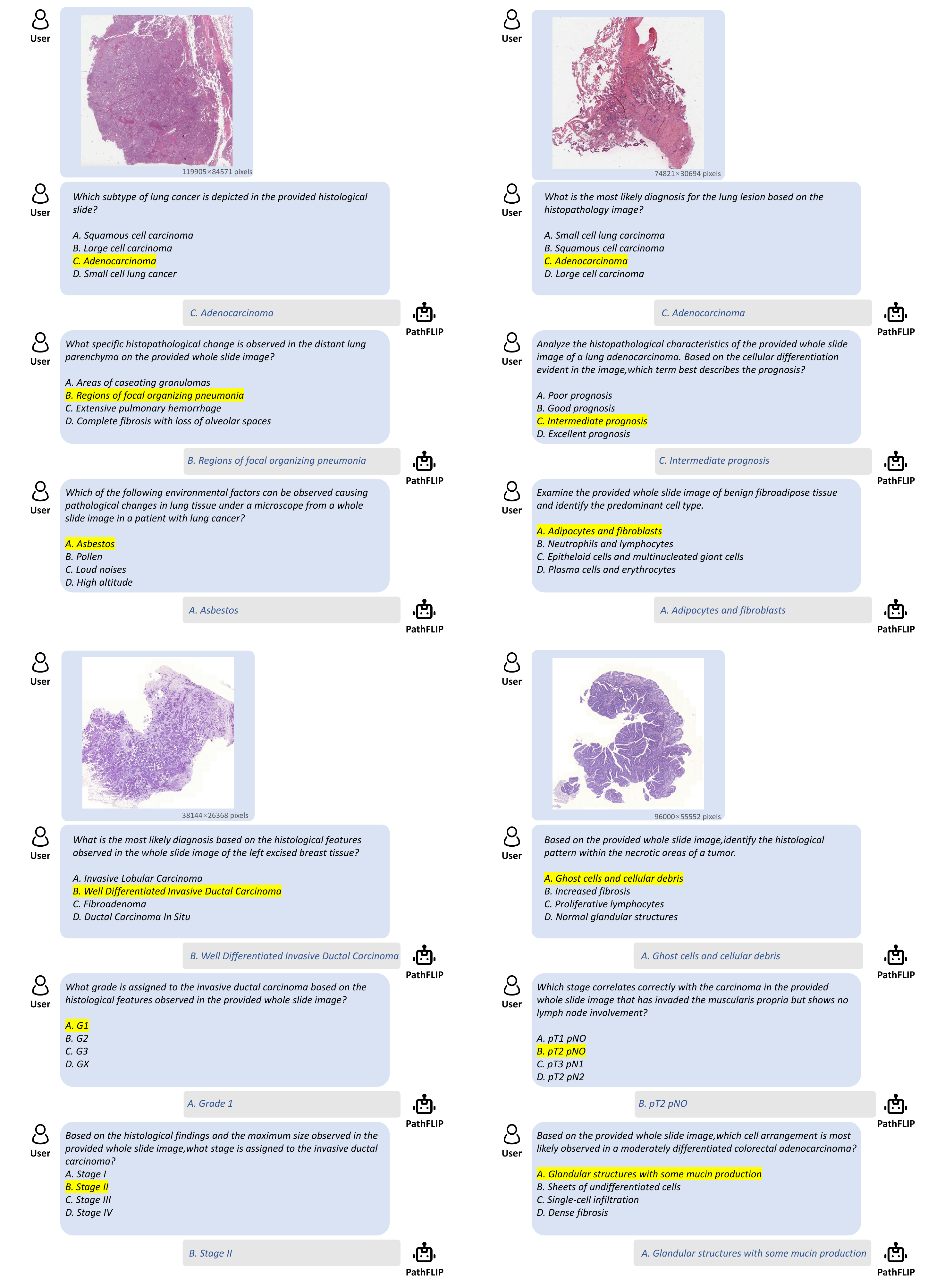}
\caption{\textbf{Results on VQA tasks.} The figure shows example input slides, corresponding questions, and the model's predicted answers. The correct answers are highlighted in yellow.}
\label{vqa2}
\end{figure*}

\subsubsection{WSI Caption and Visual Question Answering}
In Figure~\ref{caption}, we present additional examples of captions generated by our method. The results demonstrate that our model effectively captures fine-grained, region-level morphological features, such as lymphocytic infiltration and tumor grading, thanks to its region-aware captioning capability. 
Compared to PathGen, which creates slide-level captions by aggregating patch-level predictions, our approach generates descriptions that better mirror pathologists' diagnostic reasoning, including staging, grading, and region-level assessments.
In the SlideBench experiments, our model performed well on the closed-set VQA task (Figure~\ref{vqa2}), providing accurate answers to domain-specific queries, including tasks such as morphological analysis, subtype classification, staging, grading, and risk assessment.

These results indicate that our method generates clinically relevant slide-level captions with improved logical coherence, enhancing interpretability and clinical utility. Furthermore, its strong performance across multiple downstream tasks suggests potential as a diagnostic assistant for pathologists, advancing AI applications in computational pathology.

\end{document}